\documentclass[journal]{IEEEtran}
\usepackage{amsmath,amsfonts}
\usepackage{algorithm}
\usepackage{algpseudocode}  
\usepackage{booktabs}       
\usepackage{bm}             
\usepackage{array}
\usepackage[caption=false,font=normalsize,labelfont=sf,textfont=sf]{subfig}
\usepackage{textcomp}
\usepackage{stfloats}
\usepackage{url}
\usepackage{verbatim}
\usepackage{graphicx}
\usepackage{cite}
\usepackage{bbding}
\usepackage{CJKutf8}

\usepackage[colorlinks,linkcolor=red,citecolor=blue]{hyperref}

\usepackage{xcolor}  
\usepackage{bbding}   
\usepackage{multirow}  
\usepackage{makecell} 
\usepackage{tikz}  
\newcommand{\circled}[1]{
    \tikz[baseline=(char.base)]{
        \node[shape=circle, draw, inner sep=0pt, outer sep=0pt, minimum size=1pt] (char) {#1};
    }
}

\usepackage{textcomp}  
\newcommand{\tb}{\textbf}
\newcommand{\et}[2]{${#1}^{\pm{#2}}$} 
\newcommand{\ets}[2]{$\underline{{#1}}^{\pm{#2}}$} 
\newcommand{\etb}[2]{$\mathbf{{#1}}^{\pm{#2}}$} 
\newcommand{\pub}[1]{\color{gray}{\tiny{#1}}} 

\begin{document}
\begin{CJK}{UTF8}{gbsn}

\title{Coordinating Multiple Conditions for Trajectory-Controlled Human Motion Generation}

\author{Deli Cai, Haoyang Ma, Changxing Ding, \textit{Senior Member, IEEE}

\thanks{Deli Cai and Haoyang Ma are with the School of Electronic and Information Engineering, South China University of Technology, 381 Wushan Road, Tianhe District, Guangzhou 510000, P.R. China (e-mail: eecaideli@mail.scut.edu.cn, eemahaoyang@mail.scut.edu.cn).}

\thanks{Changxing Ding is with the School of Electronic and Information Engineering, South China University of Technology, 381 Wushan Road, Tianhe District, Guangzhou 510000, P.R. China, and also with the Pazhou Lab, Guangzhou 510330, China (e-mail: chxding@scut.edu.cn).}
}

\markboth{IEEE Transactions on Multimedia, 2026}%
{Shell \MakeLowercase{\textit{et al.}}: A Sample Article Using IEEEtran.cls for IEEE Journals}


\maketitle

\begin{abstract}

Trajectory-controlled human motion generation aims to synthesize realistic human motions conditioned on both textual descriptions and spatial trajectories. However, existing methods suffer from two critical limitations: first, the conflict between text and trajectory conditions disrupts the denoising process, resulting in compromised motion quality or inaccurate trajectory following; second, the use of redundant motion representations introduces inconsistencies between motion components, leading to instability during trajectory control.
To address these challenges, we propose CMC, a decoupled framework that effectively coordinates text and trajectory conditions through a divide-and-conquer strategy. CMC follows a divide-and-conquer paradigm, comprising two cascaded stages: Trajectory Control and Motion Completion. In the first stage, a diffusion model generates a simplified representation of the controlled joints under trajectory guidance, based on the given trajectories, ensuring accurate and stable trajectory following. In the second stage, a text-conditioned diffusion inpainting model generates full-body motions using the simplified representation from the first stage as partial observations. To mitigate overfitting caused by limited inpainting training data, we further introduce the Selective Inpainting Mechanism (SIM), which alternates between text-to-motion generation and motion inpainting tasks during training.
Experiments on HumanML3D and KIT datasets demonstrate that CMC achieves state-of-the-art performance in control accuracy and motion quality, demonstrating its effectiveness in coordinating multimodal conditions and representations. Project page: \url{https://cdlchoi.github.io/cmc_page}

\end{abstract}

\section{Introduction}
\label{sec:intro}

Trajectory-controlled human motion generation \cite{mdm,zhang2022motiondiffuse,t2mgpt,mld} aims to synthesize realistic human motions conditioned on both textual descriptions and spatial trajectories. This technology has significant potential applications in fields such as animation \cite{2022AvatarCLIP}, gaming \cite{holden2017pfnn}, and robotics \cite{2020Long}. However, it still presents a challenge due to the inherent disparity in information granularity between text and trajectory data. Specifically, language is typically abstract and conveys coarse-grained information, whereas trajectory data is significantly more detailed and fine-grained.

To improve controllability in motion generation, recent diffusion-based methods \cite{mdm, priormdm, gmd, xie2023omnicontrol} have proposed pre-assigning the spatial trajectories of specific human joints during motion generation, which can be divided into inpainting-based methods and guidance-based methods.
Inpainting-based methods \cite{mdm, priormdm} use a diffusion model to complete a full-body motion representation using the local positions of controlled joints. However, the practicality is relatively poor because the local positions are typically unknown \cite{xie2023omnicontrol}.
To address this issue, guidance-based methods \cite{gmd, xie2023omnicontrol} integrate the trajectory conditions into each denoising step in a classifier-guidance manner \cite{classifierguidance}, where trajectory guidance and text-based denoising are applied alternatively.

\begin{figure}[t]
    \vspace{0mm}
    \centering
    \includegraphics[width=1.0\linewidth]{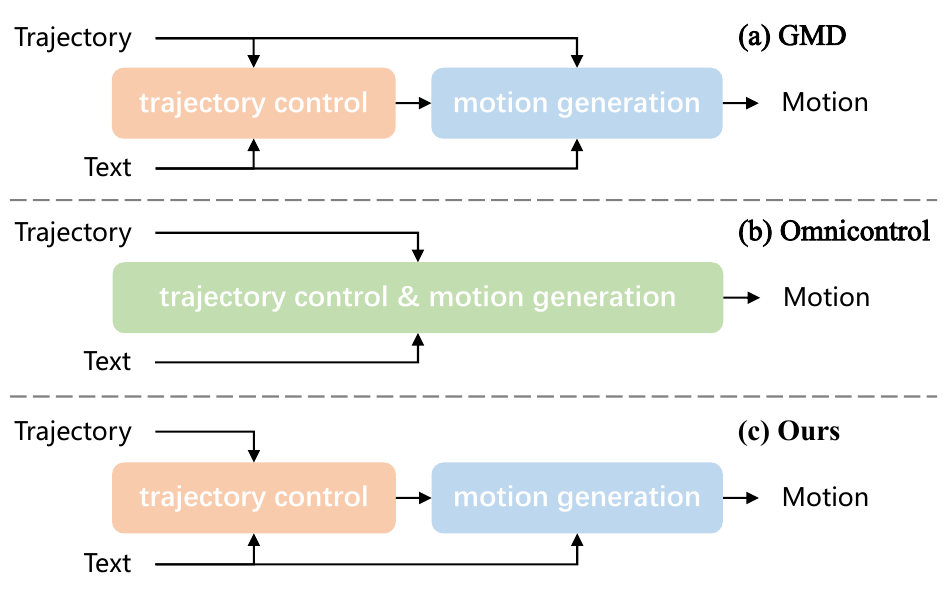}
    \vspace{-5mm}
   \caption{
        Comparison of frameworks between our approach and two mainstream frameworks. (a) GMD \cite{gmd} applies both text and trajectory conditions at both stages, particularly during the motion generation stage. (b) Omnicontrol \cite{xie2023omnicontrol} generates motion in a single integrated stage, utilizing both text and trajectory conditions simultaneously. (c) In contrast, our CMC decouples the trajectory control and motion generation stages, thereby avoiding conflicts between the text and trajectory conditions.
        }
        
    \label{fig:3_framework}
\end{figure}

\begin{figure}[t]
    \vspace{0mm}
    \centering
    \includegraphics[width=0.9\linewidth]{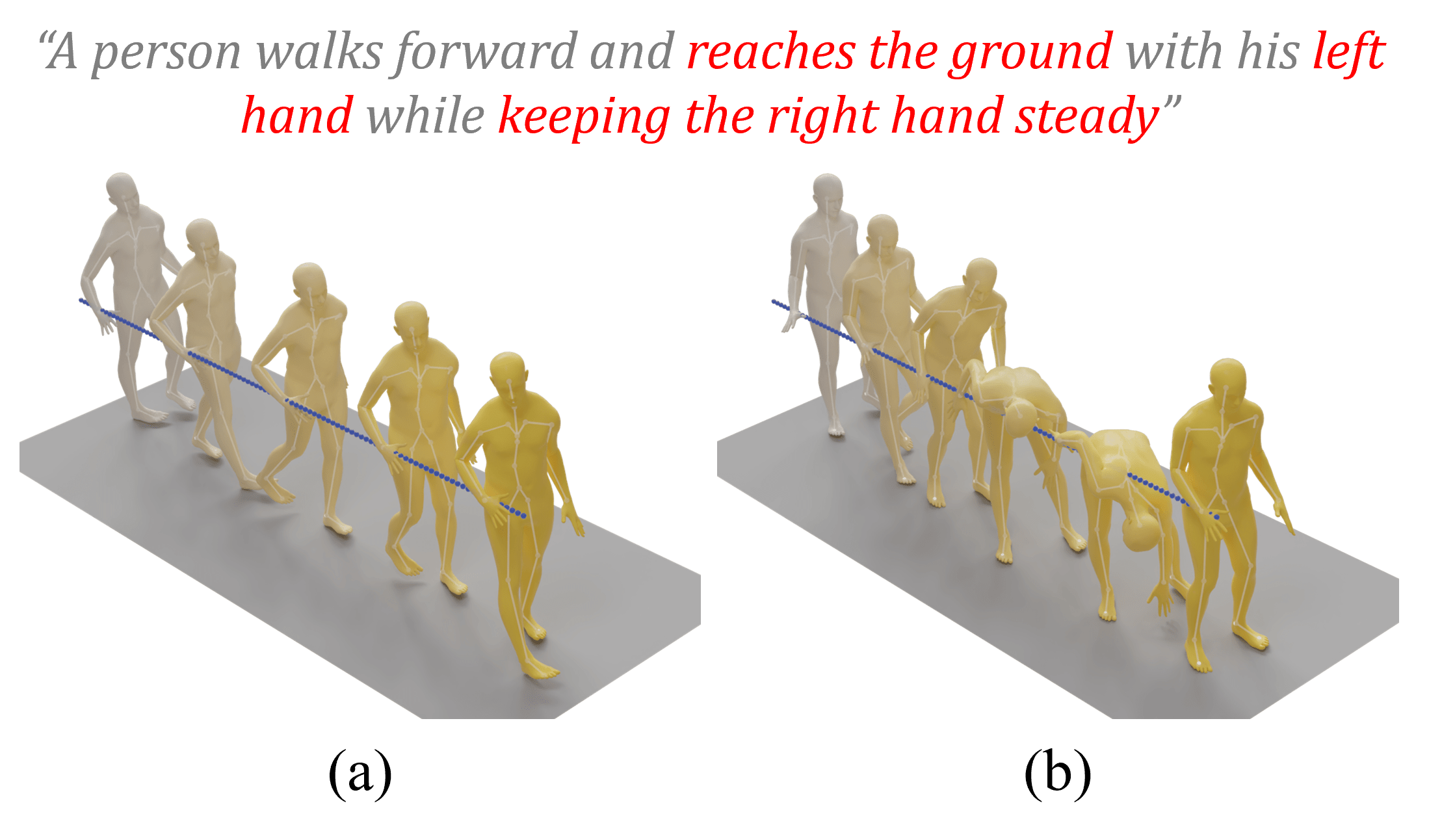}
    \vspace{-3mm}
   \caption{
        (a) With conflict: insufficient understanding of the text condition leads to suboptimal motion quality. (b) Without conflict: the network fully maps the text to motion while achieves accurate trajectory following.
        }

    \label{fig:intro_comp_img}
\end{figure}

\begin{figure}[t]
    \vspace{0mm}
    \centering
    \includegraphics[width=0.9\linewidth]{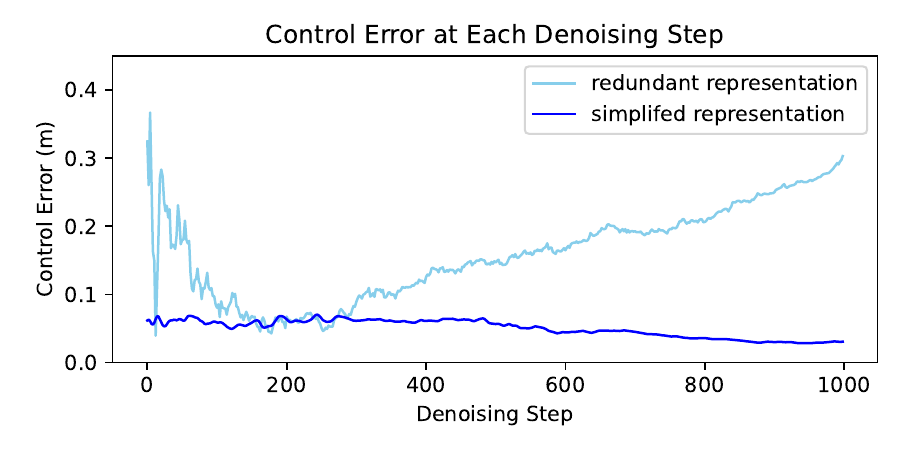}
    \vspace{-3mm}
   \caption{
        Comparisons of the control error for the redundant and simplified representations across all denoising steps.
        }

    \label{fig:intro_error}
\end{figure}

However, we identify several fundamental limitations in prior works~\cite{gmd,xie2023omnicontrol} that follow the paradigm of simultaneous trajectory control and motion generation (Fig.~\ref{fig:3_framework}). The primary issue lies in the conflict between text and trajectory conditions and the inconsistencies introduced by redundant motion representations.
We argue that injecting the trajectory guidance into the diffusion process disrupts the original denoising process. This interference may hinder the network from following the learned text-to-motion mapping during training, as illustrated in Fig.~\ref{fig:intro_comp_img}. As a result, the generated motion can become semantically inconsistent with the input text.
Moreover, using the redundant representation (including local joint positions, rotations, and velocities) \cite{gmd, xie2023omnicontrol} for trajectory-controlled motion generation may undermine control accuracy and lead to instability during trajectory control, as illustrated in Fig.~\ref{fig:intro_error}.
However, trajectory guidance is usually applied only to the local positions, which leads the inconsistency between the components in redundant representation \cite{meng2024mardm} and leaves the diffusion network to implicitly reconcile inconsistencies among these components during denoising. 
Motivated by these observations, we explore coordinating the multiple
conditions and representations during motion generation

\begin{figure*}[t!]
    \vspace{0mm}
    \centering
       \includegraphics[width=1.0\linewidth]{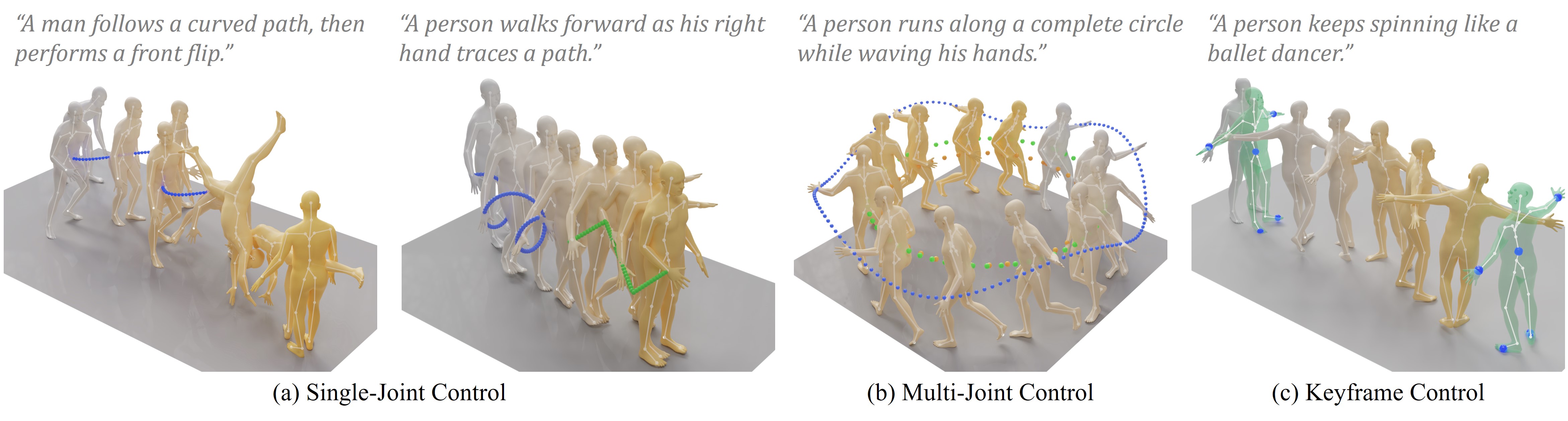}
       \vspace{-4mm}
        \caption{We propose to Coordinate Multiple Conditions (CMC) for trajectory-controlled human motion generation. We visualize examples using different conditions: (a) Single-Joint Control: the specified trajectory for one joint covers partial or all frames. (b) Multi-Joint Control: the specified trajectories for multiple joints are available. (c) Keyframe control: joint locations are available only on several keyframes. We visualize the human body while keeping its joints visible to demonstrate the high controllability of our CMC.
            }
       \label{fig:teaser_image}
    \vspace{0mm}
\end{figure*}

Hence, we propose to coordinate multiple conditions (CMC) for trajectory-controlled human motion generation. CMC is a decoupled framework that leverages text and trajectory conditions separately. It follows a divide-and-conquer paradigm, comprising two cascaded stages: Trajectory Control and Motion Completion, based on diffusion models. 
In the Trajectory Control stage, a diffusion model first generates plausible local joint positions for the pelvis and controlled joints conditioned on the input text, which we denote as a \textit{simplified representation}. Following the diffusion model, we apply trajectory guidance in a classifier-guidance manner to refine this simplified representation according to the given trajectories.
This simplified representation makes the control process more stable and results in better control accuracy. Once optimized, the positions of the controlled joints are fixed and passed to the next stage.

The Motion Completion stage is also built on a diffusion model and leverages only the text condition to produce the full-body motions \cite{humanml3d, mld, t2mgpt}. It utilizes the simplified representation from the first stage (i.e., the local positions of the pelvis and the controlled joints) as partial observations and generates full-body motion using a diffusion inpainting model \cite{mdm}. Crucially, these trajectory-relative components remain frozen throughout this stage, thereby preserving the refined spatial path.
Further, we observe that training the diffusion inpainting model exclusively on the inpainting task leads to overfitting, primarily due to the limited size of the motion inpainting data. To address this issue, we propose a Selective Inpainting Mechanism (SIM) for training, which dynamically alternates between two training modes: (1) standard text-to-motion generation, and (2) motion inpainting. The former task serves as a regularization approach, enhancing the generalization of the inpainting model while maintaining its inpainting capability.

Our CMC can perform various tasks, such as single-joint control, multi-joint control, and keyframe control, as illustrated in Fig. \ref{fig:teaser_image}. We emphasize that our CMC addresses the conflict between text and trajectory conditions. In the first stage, the use of a simplified representation minimizes the influence of text on trajectory control, enabling stable and accurate alignment with the desired path. This improvement is clearly reflected in the reduced and stable error observed in Fig.~\ref{fig:intro_error}. In addition, under the decoupled paradigm, the second stage generates full-body motion solely from the text prompt \textit{without} trajectory guidance, thereby preserving the integrity of the original denoising process and avoiding interference with semantic motion generation, as illustrated in Fig. \ref{fig:intro_error} (b).

Overall, our contributions can be summarized as follows: 
\begin{enumerate}
    \item We identify two key challenges in current text and trajectory-controlled human motion generation: the conflict between text and trajectory conditions, and the inconsistencies introduced by redundant motion representations.
    \item We propose a decouple framework named CMC to coordinate the text and trajectory conditions. Our framework leverages the two conditions separately and significantly alleviates the conflict between multiple conditions and motion representations.
    \item To improve the generalization of diffusion inpainting model, we propose to apply the selective inpainting mechanism (SIM) during the training. We show that a plug-and-play SIM significantly improves the generalization of the diffusion inpainting model.
    \item Our framework achieves state-of-the-art control accuracy and motion quality on the popular HumanML3D and KIT datasets, demonstrating its effectiveness.
\end{enumerate}

\section{Related Works} \label{sec:related_works}

\subsection{Single-Conditional Human Motion Generation}

Single-conditional human motion generation aims to synthesize realistic and diverse motion sequences aligned to the given condition. According to the different modalities of the condition, single-conditional human motion generation methods can be roughly divided into text-to-motion generation \cite{mdm, mld, guo2023momask, t2mgpt}, audio-to-motion condition generation \cite{emotiongesture_tmm,deepdance_tmm,probtalk,icgn_tmm}, and motion-to-motion generation (conditioned on historical motion or partner's motion, e.g., motion prediction and reaction generation) \cite{divdiff_tmm, InterFormer_tmm, manet_tmm}. The most relevant to our work is text-to-motion generation. 

Text-to-motion generation produces human motion sequences based on textual descriptions. 
Early text-to-motion generation methods focused on aligning the latent space between textual descriptions and motion \cite{petrovich2022temos, petrovich2023tmr}. However, due to the inherent disparity of the distributions between text and motion, forcibly aligning their distributions is challenging \cite{tevet2022motionclip}.

According to the generative framework, mainstream text-to-motion generation methods can be primarily categorized into vector quantized variational autoencoder (VQVAE) and diffusion-based methods. 
The VQVAE method \cite{vqvae}, an extension of VAE, learns discrete latent spaces instead of continuous ones using the vector quantization technique. VQVAE-based methods \cite{humanml3d, t2mgpt, jiang2024motiongpt, guo2023momask, pinyoanuntapong2023mmm} use a token prediction network to map the text condition into the categorical distribution of tokens. Then, they map the predicted tokens into a motion sequence using a decoder.
According to the mode of token prediction, existing approaches can be classified into two categories: (1) autoregressive prediction methods \cite{t2mgpt, jiang2024motiongpt} and (2) masked modeling methods \cite{guo2023momask, pinyoanuntapong2023mmm}.
Autoregressive prediction methods predict the next token based on the previously predicted ones, using text conditions to generate the motion sequence.
Instead of focusing only on previous tokens, the masked modeling method examines the contextual cues of the motion sequence and predicts multiple tokens during each iteration. Consequently, masked modeling significantly reduces generation time.
Furthermore, many methods \cite{scamo, guo2023momask} explore adopting advanced quantizers to achieve better performance.

Diffusion-based methods \cite{DDPM} model human motion distributions by denoising the noisy input. Based on the representation space where the diffusion process occurs, diffusion-based methods can be categorized into standard \cite{mdm, zhang2022motiondiffuse, kim2023flame, he2023semanticboost} and latent diffusion-based \cite{mld, jin2024graphmotion} strategies. 
Standard diffusion-based methods use transformer-based backbone networks to perform the denoising process on the raw motion space with additional conditional embedding.
In contrast, latent diffusion-based methods map motion sequences from raw motion space to a low-dimensional motion latent space using a variational autoencoder. Then, they perform the diffusion process on the latent space \cite{mld, jin2024graphmotion}.

Although text-to-motion generation methods have successfully generated realistic motion sequences, they struggle to control human joints accurately. In this paper, we aim to coordinate the incorporation of trajectory conditions into the text-to-motion generation framework.

\vspace{1mm}
\subsection{Multi-Conditional Human Motion Generation}
Due to human motion's inherent complexity and diversity, exploring its generation according to multiple conditions is highly meaningful. Depending on the condition compositions, current multi-conditional motion generation methods primarily include trajectory-controlled motion (i.e., text and trajectory) \cite{pinyoanuntapong2025maskcontrol,crowdmogen, gmd, xie2023omnicontrol, wan2023tlcontrol, gu2026bridging}, human-object interactions (i.e., text and object) \cite{ghosh2023imos, peng2023hoidiff, diller2024cghoi, mengqing}, human-scene interactions (i.e., text and scene) \cite{wang2022humanise, yi2025tesmo}, motion style transfer (i.e., text and motion) \cite{guo2024style, zhong2024smoodi, 2024guidedmotion}, speech motion generation (i.e., text and audio) \cite{liu2022beat, dabral2023mofusion, yoon2020speech, emotiongesture_tmm, hop}, and sketch to motion \cite{wang2025stickmotion}. 
The methods used for fusing multiple conditions can be classified into two categories: (1) implicit fusion, which integrates the embeddings of multiple conditions using attention mechanisms \cite{transformer} or ControlNet \cite{controlnet}; and (2) explicit guidance, which employs a task-specific objective function \cite{classifierguidance, diller2024cghoi} during or after the generation process \cite{wan2023tlcontrol}. In trajectory-controlled motion generation, explicit guidance is widely used by mainstream methods \cite{gmd, xie2023omnicontrol, wan2023tlcontrol} due to the trajectory condition's high precision requirement.
However, these methods \cite{gmd, xie2023omnicontrol} apply text and trajectory conditions using implicit fusion and explicit guidance in an interleaved manner, which leads to instability during the generation process and results in suboptimal motion quality. To address this issue, we propose decoupling motion control from text-conditioned motion generation to enhance the coordination of text and trajectory conditions during trajectory-controlled human motion generation.

\begin{figure*}[t] 
    \vspace{0mm}
    \centering
       \includegraphics[width=1.0\linewidth]{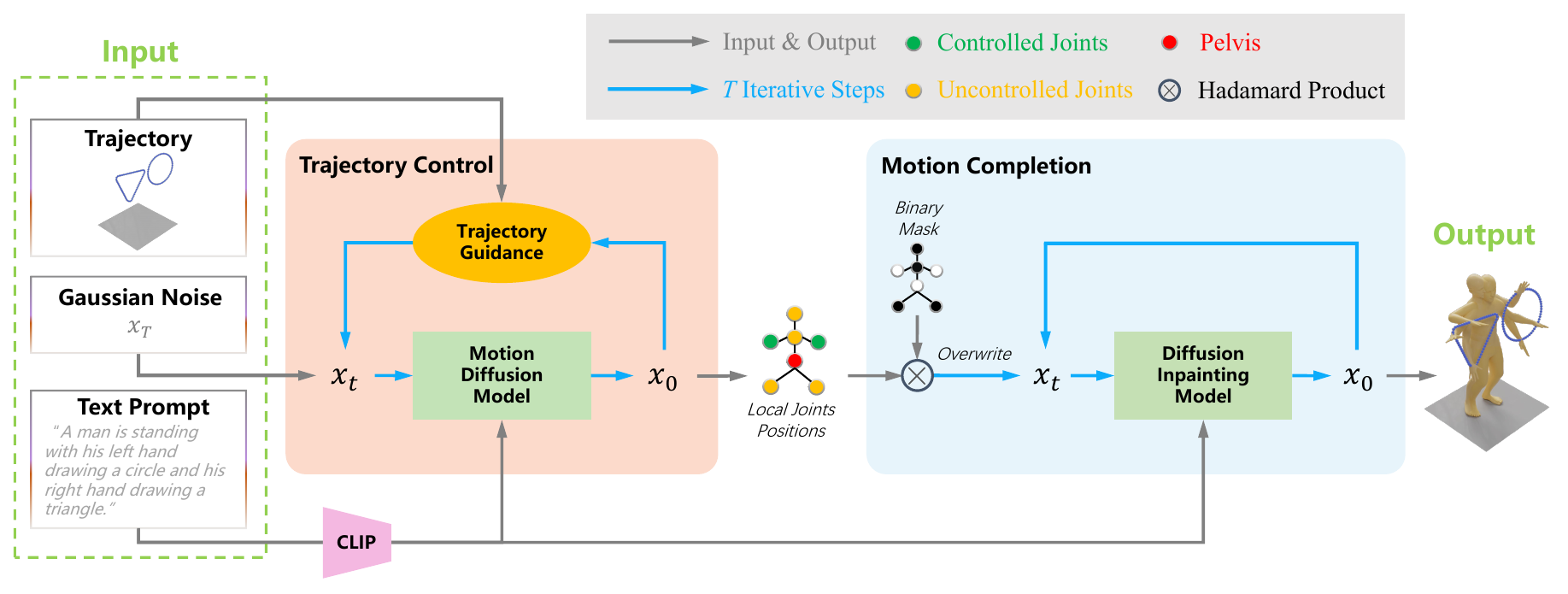}
       \vspace{-5mm}
       \caption{
            \textbf{Overview of our CMC. It consists of two stages: Trajectory Control and Motion Completion.} In the \textit{Trajectory Control} stage, we utilize textual descriptions and spatial trajectories of the controlled joints to predict the trajectories of both the pelvis and the controlled joints within a simplified representation space. Subsequently, the \textit{Motion Completion} stage takes these trajectories as partial observations and completes the full-body motion using a diffusion inpainting model. Both stages use CLIP as the text encoder.
            }
    \vspace{-1mm}
    \label{fig:overview}
\end{figure*}

\section{Methodology}

Our proposed framework CMC generates human motion sequences that semantically align with the text prompts while ensuring that the joints closely adhere to the given spatial trajectories. CMC employs a divide-and-conquer strategy, utilizing text and trajectory conditions separately and consisting of two stages: \textit{Trajectory Control} and \textit{Motion Completion}. 
In the first stage, the text prompt and the given spatial trajectories of the controlled joints are used to generate local joint positions of the pelvis and controlled joints (a simplified motion representation) (\ref{sec:stage1}). At this point, the trajectory control process is complete, and the controlled joints are fully determined. 
Subsequently, in the second stage, these local positions are used as partial observations to synthesize the final full-body motion, conditioned solely on the input text prompt (\ref{sec:stage2}).

\subsection{Preliminary} \label{sec:preliminary}

The vanilla diffusion-based trajectory-controlled motion generation framework includes two key components: the denoising model and trajectory guidance. Given text prompt \textit{P}, spatial trajectory condition \textit{C} and the Gaussian noise $\mathbf{x}_T$, this framework's denoising function can be formulated as $M(\hat{ \mathbf{x}_0}|\mathbf{x}_{T}, \textit{P}, \textit{C})$. 

\vspace{2mm}
\textit{Diffusion process.} Starting with a motion sequence $\mathbf{x}_0$ sampled from the training data distribution, the diffusion process transforms $\mathbf{x}_0$ into noisy data $\{\mathbf{x}_t\}_{t=1}^{T}$ with varying noise levels, where $T$ is the maximum diffusion process step. This process is formulated as  $\mathbf{x}_t=\sqrt{1-\beta_t}\mathbf{x}_{t-1}+\sqrt{\beta_t}\epsilon$, where $\beta_t$ is the noise scheduler and $\epsilon$ is gaussian noise. 
A notable property of the forward process is that it admits sampling $\mathbf{x}_t$ at an arbitrary timestep $t$ in one step:
\begin{equation} \label{equ:one_step_diffusion}
    \mathbf{x}_t=\sqrt{1-\beta_t}\mathbf{x}_0+\sqrt{\beta_t} \epsilon=\boldsymbol{\alpha}_t \mathbf{x}_0 + \boldsymbol{\sigma}_t \epsilon.
\end{equation}

\noindent Thus, $\mathbf{x}_t$ can be viewed as the weighted sum of $\mathbf{x}_0$ and $\epsilon$. We denote their coefficient as $\boldsymbol{\alpha}_t$ and $\boldsymbol{\sigma}_t$ for brevity, and $\boldsymbol{\alpha}_t ^2 +  \boldsymbol{\sigma}_t ^2=1$.

\vspace{2mm}
\textit{Denoising process.} The denoising process models the conditional distribution $p_\theta(\hat{\mathbf{x}_0}|\mathbf{x}_T,\textit{P})$ based on the text prompt \textit{P} and obtains the synthesized motion $\hat{\mathbf{x}_0}$. This was achieved using a denoising network to denoise $\mathbf{x}_t$ to $\hat{\mathbf{x}_0}$, changing $t$ from $T$ to 1.

To ensure that the synthesized motion closely follows the pre-assigned trajectory, trajectory guidance is applied after each denoising step. 
First, the posterior mean $\mu_t$ was computed with a function $\mu_t=F(\hat{\mathbf{x}_0},\mathbf{x}_t)$ following \cite{DDPM}. Similar to classifier guidance \cite{classifierguidance}, trajectory guidance uses an optimizer to refine the posterior mean $\mu_t$ at each denoising step by minimizing the objective function $G$ between the synthesized motion $\hat{\mathbf{x}_0}$ and the spatial trajectory condition \textit{C}. 
Since the pre-assigned trajectory is defined in the three-dimensional world coordinate system and the motion within the network is represented locally, a conversion function $R$  was used to transform the synthesized motion from the local to the global coordinate system. Finally, the gradient from the objective function $G$  was backpropagated to $\mu_t$ and updates the motion. This process can be expressed by the following formula: $\mu_t = \mu_t - \tau \nabla_{\mu_t} G(R(\mu_t),\textit{C})$, where $\tau$ controls the guidance strength. The most related work, GMD\cite{gmd} and Omnicontrol\cite{xie2023omnicontrol}, follow this vanilla paradigm.

Although the vanilla diffusion-based trajectory-controlled motion generation framework can achieve motion control to a certain extent, the interleaved manner of applying text and trajectory conditions may lead to suboptimal motion quality. Hence, we introduce our proposed decoupled CMC to tackle this challenge, as shown in Fig. \ref{fig:overview}.

\textit{Redundant motion representation.} The widely-adopted redundant motion representation, proposed by Guo et al. \cite{humanml3d}, is defined by a tuple of ($r^a, r^x, r^z, r^y, \mathbf{j}^p, \mathbf{j}^r, \mathbf{j}^v, \mathbf{c}^f $), 
where $r^a \in \mathbb{R}$ is the root angular velocity along the Y-axis; ($r^x, r^z \in \mathbb{R}$) are root linear velocities on the XZ-plane (i.e., horizontal plane); 
$r^y \in \mathbb{R}$ is root height, 
and $\mathbf{j}^p \in \mathbb{R}^{3j}$, $\mathbf{j}^r \in \mathbb{R}^{6j}$, $\mathbf{j}^v \in \mathbb{R}^{3j}$ is the local \underline{p}ositions, \underline{r}otations, and \underline{v}elocities of joints, with $j$ denoting the number of joints.  
$\mathbf{c}^f \in \mathbb{R}^4$ is the foot contact label, consisting of four binary labels obtained by thresholding the heel and toe joint velocities to emphasize the foot ground contacts.

\subsection{Trajectory Control} \label{sec:stage1}

The Trajectory Control stage produces local positions of the pelvis and the controlled joints, based on the textual descriptions and pre-assigned spatial trajectory of the controlled joints. This stage is built upon the Motion Diffusion Model (MDM) \cite{mdm}. Unlike previous methods \cite{gmd, xie2023omnicontrol} that employ redundant motion representations \cite{humanml3d}, we propose using a simplified representation that only includes the local joint positions in this stage. The analyses are as follows. 

We visualize a dynamic process of control error during trajectory guidance when using redundant motion representation, as employed by GMD \cite{gmd} and Omnicontrol \cite{xie2023omnicontrol}, as depicted by the light red curve in Fig. \ref{fig:1000}. The light red curve denotes $\hat{\mathbf{x}_0}$ and demonstrates evident instability, with the control error tending to increase towards the end. This phenomenon is attributed to the use of redundant motion representation.
In diffusion-based trajectory-controlled motion generation, the objective is to obtain a motion sequence where the local joint positions conform to the given trajectory. When using the redundant representation, the denoising network implicitly models the consistency between the three components during training \cite{meng2024mardm}. However, only the local positions are guided by the trajectory condition during the denoising process. The guided local joint positions achieve better control accuracy after guidance, while the rotation and velocity remain unchanged, leading to inconsistencies between the three components. These unchanged components correspond to the control accuracy before refinement. Consequently, the diffusion model attempts to compensate for these inconsistencies at the next denoising step, making the trajectory control process unstable and ultimately degrading the control accuracy.

Therefore, to achieve a more stable control process and better control accuracy, we use the simplified motion representation that only includes the local joints' positions in this stage. Our proposed simplified representation discards joints’ rotations, velocities, and foot contact relative to the redundant representation introduced in Section \ref{sec:preliminary}. Specifically, the simplified representation is denoted as a tuple of ($r^a, r^x, r^z, r^y, \mathbf{j} ^p$).

\subsection{Motion Completion} \label{sec:stage2}

The Motion Completion stage is also built with diffusion model. It uses  the text condition and the trajectories of the pelvis and controlled joints (i.e., the simplified representation) as partial observations to complete the full-body motion representation. In this stage, we use the popular redundant motion representation proposed by \cite{humanml3d} as the final output.
We adopt the diffusion inpainting technique \cite{mdm}, in which the diffusion model produces the missing components from the simplified representation.

During the generation process, we take the local positions of the pelvis and control joints from the first stage as partial observations. This operation minimizes the conflicts between text and trajectory conditions. We substitute them into the input noisy motion tensors with a binary spatial-temporal mask at each denoising step, ensuring that the controlled joints' accuracy remains fixed throughout this stage. 
In addition, the substitution operation serves as a special guidance mechanism, spreading the partial observation features to the entire motion tensor, thereby ensuring that the predicted part are coherent with the given observations. 
Overall, there is no need to focus on trajectory control during this stage, and the controlled joints remain fixed. Instead, text understanding and motion completion should be the primary focus, enabling high-fidelity motion generation. Moreover, this design eliminates the conflicts between the text and trajectory conditions, further improving the motion quality.

Additionally, we propose a simple yet effective selective inpainting mechanism (SIM) tailored for the diffusion inpainting model to enhance its generalization.
The inspiration for SIM is as follows. Language drift phenomenon has been observed in language models \cite{language_drift}, where a model pre-trained on a large text corpus and later finetuned for a specific task progressively loses syntactic and semantic knowledge of the language. A similar phenomenon of prior forgetting also exists in the diffusion model \cite{ruiz2023dreambooth, mvdream}. When a large pretrained diffusion model is finetuned on a subject-driven image generation dataset, it may forget its pretrained prior for general image generation. Moreover, there is a risk of reducing the diversity of the generated images. In the case of motion completion, we observe that training the diffusion inpainting model exclusively on inpainting tasks leads it to overfit to partial observations. The model becomes overly reliant on partial observations and loses its ability to generate diverse, natural motions when presented with unseen inputs.

\begin{figure}[t]
    \vspace{0mm}
  \centering
    \includegraphics[width=\linewidth]{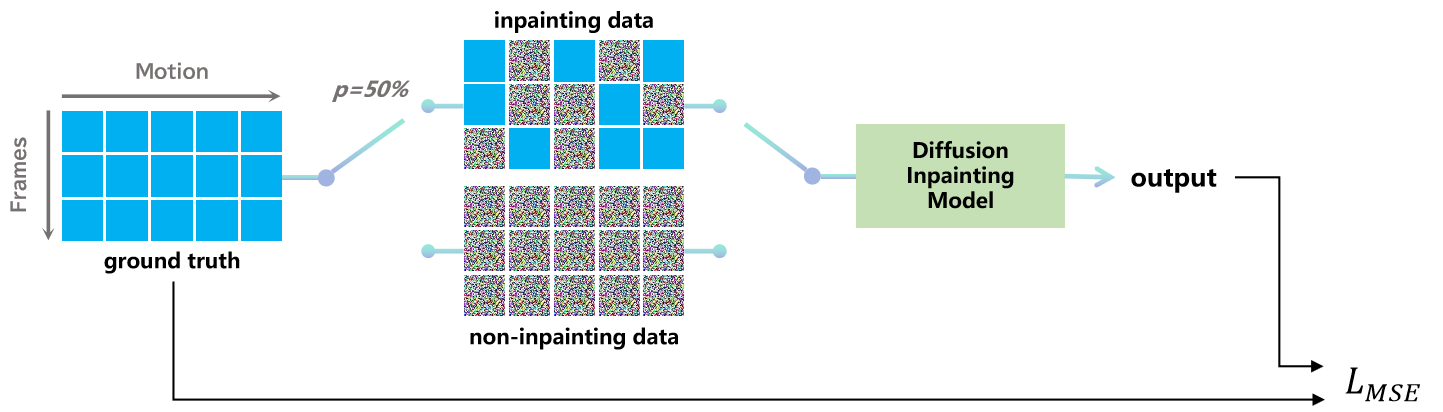}
    
    \caption{
    The workflow of SIM to train the diffusion inpainting model. SIM prepares non-inpainting data and inpainting data with a probability of 50\% respectively.
    }
    
    \vspace{0mm}
    \label{fig:sim_flow}
\end{figure}

\begin{figure}[t]
    \vspace{0mm}
  \centering
    \includegraphics[width=0.9\linewidth]{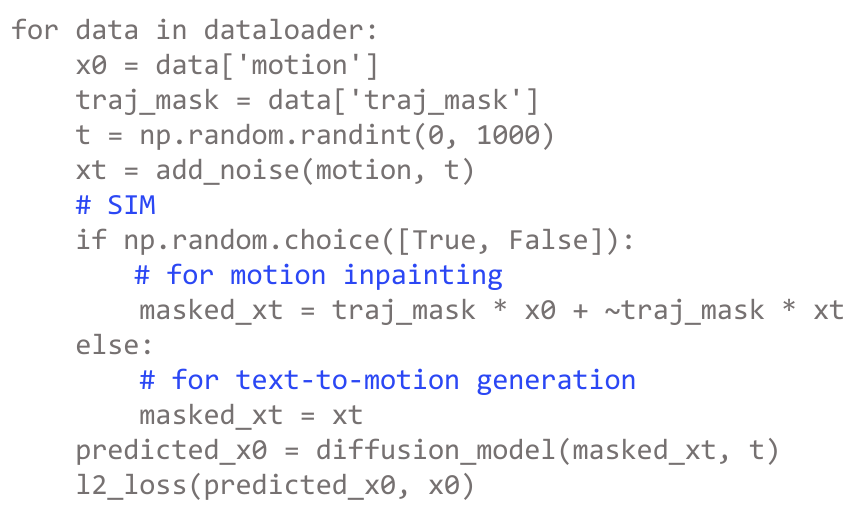}
    \vspace{-4mm}
    \caption{
    Pseudo-code of our SIM implementation in Python.
    }
    \vspace{-4mm}
    \label{fig:pseudo_code_sim}
\end{figure}

The workflow of SIM is illustrated in Fig. \ref{fig:sim_flow}.
Specifically, SIM prepares two types of motion data for training: non-inpainting and inpainting. Non-inpainting data includes features that are mixed with noise, while inpainting data provides the necessary partial observations for motion completion (i.e., a portion of clean data). The two types of data are used with a probability of 50\% respectively. SIM serves two primary purposes. First, training with inpainting data allows the model to generate full-body motion from partial observations. Second, training with non-inpainting data functions as a regularization technique, reducing overfitting and improving the model's generalization ability to unseen partial observations. We also presented the pseudo code of our SIM implementation in Fig. \ref{fig:pseudo_code_sim}. We demonstrate the effectiveness of SIM under the motion inpainting setting and the text-to-motion generation setting in Sections \ref{subsec:ablation_component} and \ref{sec:sim}.

\vspace{-2mm}
\subsection{Training Target} \label{sec:loss}

We employ two loss functions to train both stages. One is the element-wise loss implemented with mean square error loss (MSE), which trains the model to denoise noisy motion into clean motion $\hat{\mathbf{x}_0}$:

\begin{equation}
\mathcal{L}_{elem}  = \frac{1}{N_{elem}} \sum_{i=1}^{N_{elem}} \left \| \mathbf{x}_0 -  \hat{\mathbf{x}_0}\right \|  ^2_2,
\end{equation}

\noindent where $N_{elem}$ denotes the number of elements in motion tensor.

We note that relying only on an element-wise loss can lead to insufficient constraints on the global trajectory. The element-wise loss alone can result in considerable cumulative errors because motion is represented locally while the trajectory is defined in the world coordinate system. To address this issue, we introduce a global position loss that constrains the positions of the globally controlled joints. This can be expressed using the following formula:

\begin{equation}
\mathcal{L}_{global}  = \frac{1}{N_{cont}} \sum_{i=1}^{N_{cont}} \left \| R(\hat{\mathbf{x}_0}) -  \textit{C} \right \|  ^2_2,
\end{equation}

\noindent where $R(\cdot)$ denotes the conversion function that converts local positions to global coordinates, \textit{C} is the given spatial trajectory, and $N_{cont}$ represents the number of all the controlled joints. Overall, our training loss is :
\begin{equation}
\mathcal{L}  = \mathcal{L}_{elem} + \mathcal{L}_{global}.
\end{equation}

\begin{table*}[t]

    \normalsize
    \centering
    \caption{
        Quantitative results on the HumanML3D \cite{humanml3d} testing set. All results are tested on the premise of controlling all frames. The best scores are highlighted in \textbf{bold}. $\rightarrow$ means closer to real data is better.
    }
    \resizebox{0.95\textwidth}{!}{
    \begin{tabular}{c|c|ccccccc} 
    \toprule
    \textbf{Controlled joint}
    & \textbf{Method }
    & \textbf{FID} $\downarrow$ 
    & \multicolumn{1}{p{2.1cm}}{\centering \textbf{R-precision} $\uparrow$ \\ \textbf{(Top-3)}} 
    & \textbf{Diversity $\rightarrow$} 
    & \multicolumn{1}{p{2.1cm}}{\centering \textbf{Foot skating} \\ \textbf{ratio} $\downarrow$ }  
    & \multicolumn{1}{p{1.8cm}}{\centering \textbf{Traj. err.} $\downarrow$  \textbf{(50 cm, \%)} } 
    & \multicolumn{1}{p{1.8cm}}{\centering \textbf{Loc. err.} $\downarrow$  \textbf{(50 cm,\%)} } 
    & \multicolumn{1}{p{1.6cm}}{\centering \textbf{Avg. err.} $\downarrow$  \textbf{(cm)}} \\
    
    \midrule
    
    - & Real & 0.002  & 0.797 & 9.503 & 0.000 & 0.000 & 0.000 & 0.000   \\ 
    
    \midrule
    \multirow{7}{*}{\makecell[c]{\textbf{Pelvis}}} 
    & MDM~\cite{mdm}~\pub{ICLR23}  & 0.698  & 0.602 & 9.197 &0.102  & 40.22& 30.76 & 59.59\\ 
    & PriorMDM~\cite{priormdm}~\pub{ICLR24} & 0.475  & 0.583 & 9.156 &0.089 & 4.57& 21.32 & 44.17\\ 
    & GMD~\cite{gmd}~\pub{ICCV23}      & 0.576  & 0.665 & 9.206 &0.101  & 9.31& 3.21& 14.39\\  
    & Omnicontrol~\cite{xie2023omnicontrol}~\pub{ICLR24} & 0.218& 0.687 & 9.422 &\textbf{0.054} & 3.87& 0.96& 3.38\\  
    & TLControl~\cite{wan2023tlcontrol}~\pub{ECCV24}      & 0.271  & 0.779 & 9.569  &-  & \textbf{0.00}& \textbf{0.00}& 1.08\\  
    & InterControl~\cite{intercontrol}~\pub{NIPS24}      & 0.159  & 0.671 & \textbf{9.482}  &0.072  & 1.32& 0.04& 4.96\\  
    & \textbf{Ours}     & \textbf{0.097}& \textbf{0.784}& 9.539 &0.068  & 0.22& 0.01& \textbf{0.51}\\
    
    \midrule
    \multirow{3}{*}{\makecell[c]{\textbf{Left foot}}} 
    & Omnicontrol~\cite{xie2023omnicontrol} & 0.280& 0.696 & \textbf{9.553} &0.069 &  5.94& 0.94& 3.14\\ 
    & TLControl~\cite{wan2023tlcontrol} & 0.368& 0.768& 9.774 &- &  \textbf{0.00}& \textbf{0.00}& 1.14\\ 
    & \textbf{Ours}   & \textbf{0.100}& \textbf{0.771}& 9.474 & \textbf{0.063}   & 1.01& 0.04& \textbf{0.85}\\ 
    
    \midrule
    \multirow{3}{*}{\makecell[c]{\textbf{Right foot}}} 
    & Omnicontrol~\cite{xie2023omnicontrol} & 0.319& 0.701& \textbf{9.481} &0.066 &  6.66& 1.20& 3.34\\ 
    & TLControl~\cite{wan2023tlcontrol} & 0.361& \textbf{0.775} & 9.778 &- &  \textbf{0.00}& \textbf{0.00}& 1.16\\ 
    & \textbf{Ours}     & \textbf{0.107}& 0.774 & 9.600 &\textbf{0.060}   & 0.95& 0.04& \textbf{0.87}\\

    \midrule
    \multirow{3}{*}{\makecell[c]{\textbf{Head}}} 
    & Omnicontrol~\cite{xie2023omnicontrol} & 0.335& 0.696& \textbf{9.480} &\textbf{0.055} &  4.22& 0.79& 3.49\\ 
    & TLControl~\cite{wan2023tlcontrol} & 0.279& 0.778& 9.606 &- &  \textbf{0.00}& \textbf{0.00}& 1.10\\ 
    & \textbf{Ours}     & \textbf{0.105}& \textbf{0.780}& 9.567 &0.068    & 0.28& 0.02& \textbf{0.75}\\

    \midrule
    \multirow{3}{*}{\makecell[c]{\textbf{Left wrist}}} 
    & Omnicontrol~\cite{xie2023omnicontrol} & 0.304& 0.680& 9.436&\textbf{0.056} & 8.01& 1.34& 5.29\\ 
    & TLControl~\cite{wan2023tlcontrol} & \textbf{0.135}& \textbf{0.789}& 9.757&-  &  \textbf{0.00}& \textbf{0.00}& 1.08\\ 
    & \textbf{Ours}    & 0.159& 0.763& \textbf{9.453} &0.068   & 0.71& 0.03& \textbf{0.87}\\

    \midrule
    \multirow{3}{*}{\makecell[c]{\textbf{Right wrist}}}
    & Omnicontrol~\cite{xie2023omnicontrol} & 0.299& 0.692& \textbf{9.519} &\textbf{0.060} & 8.13& 1.27& 5.19\\ 
    & TLControl~\cite{wan2023tlcontrol} & \textbf{0.137}& \textbf{0.787}& 9.734 &- &  \textbf{0.00}& \textbf{0.00}& 1.09\\ 
    & \textbf{Ours}      & 0.149& 0.775 & 9.627 &0.068  & 0.58& 0.03& \textbf{0.85}\\

    \midrule
    \multirow{4}{*}{\makecell[c]{\textbf{All joints above}}}
    & Omnicontrol~\cite{xie2023omnicontrol} & 2.614  & 0.606 & 8.594 &-   & 75.59& 12.30& 23.67\\ 
    & TLControl~\cite{wan2023tlcontrol} & \textbf{0.032}  & \textbf{0.794} & 9.750 &-   & \textbf{0.00}& \textbf{0.00}& 1.57\\ 
    & MotionLCM~\cite{dai2024motionlcm} & 0.444  & 0.753 & 9.355 &-   & 20.89 & 1.72 & 11.40\\ 
    & \textbf{Ours}   & 0.075& 0.783& \textbf{9.553} &\textbf{0.062} & 4.05& 0.06& \textbf{0.98}\\ 
    
    \bottomrule
    \end{tabular}
    }
    
    \vspace{-3mm}
    \label{table:humanml3d}
\end{table*}

\begin{table*}[!]

    \centering
    \scriptsize
    \caption{
        Quantitative results on the KIT-ML \cite{plappert2016kit} testing set. All data is tested on the premise of controlling all frames. The best scores are highlighted in \textbf{bold}. $\rightarrow$ means closer to real data is better. BUN means body upper neck, defined in KIT-ML.
    }
    \resizebox{0.90\textwidth}{!}{
    \begin{tabular}{c|c|cccccc} 
    \toprule
    \textbf{Controlled joint }
    & \textbf{Method}
    & \textbf{FID} $\downarrow$ 
    & \multicolumn{1}{p{2.1cm}}{\centering \textbf{R-precision} $\uparrow$ \\ \textbf{(Top-3)}} 
    & Diversity $\rightarrow$ 
    & \multicolumn{1}{p{1.6cm}}{\centering \textbf{Traj. err.} $\downarrow$ \\ \textbf{(50 cm, \%)} } 
    & \multicolumn{1}{p{1.5cm}}{\centering \textbf{Loc. err.} $\downarrow$ \\ \textbf{(50 cm, \%)} } 
    & \multicolumn{1}{p{1.6cm}}{\centering \textbf{Avg. err. }$\downarrow$ \\ \textbf{(cm)}} \\
    
    \midrule
    
    - & Real & 0.031  & 0.779 & 11.080 & 0.000 & 0.000 & 0.000   \\ 
    
    \midrule
    \multirow{5}{*}{\makecell[c]{\textbf{Pelvis}}}
    & PriorMDM & 0.851& 0.397& 10.518&  33.10& 14.00& 23.05\\ 
    & GMD      & 1.565& 0.382& 9.664&  54.43& 30.03& 40.70\\  
    & Omnicontrol & 1.417& 0.374& 10.823&  24.42& 6.25& 14.70\\
    & TLControl& 0.432& 0.757& 10.723&  \textbf{0.28}& 0.11&2.76\\  
    & \textbf{Ours}     & \textbf{0.259}& \textbf{0.772}& \textbf{11.068}&  0.57& \textbf{0.02}& \textbf{0.41}\\ 

    \midrule
    \multirow{2}{*}{\makecell[c]{\textbf{BUN}}}
    & Omnicontrol & 0.866& 0.395& 10.981&  22.16& 5.47& 13.42\\ 
    & \textbf{Ours}    & \textbf{0.245}& \textbf{0.788}& \textbf{11.114}&  \textbf{1.42}& \textbf{0.10}& \textbf{0.67}\\ 
    
    \midrule
    \multirow{2}{*}{\makecell[c]{\textbf{Left wrist}}}
    & Omnicontrol & 0.569& 0.389& 10.890&  31.39& 6.94& 18.24\\ 
    & \textbf{Ours}      & \textbf{0.312}& \textbf{0.771}& \textbf{10.985}&  \textbf{0.71}& \textbf{0.02}& \textbf{0.65}\\

    \midrule
    \multirow{2}{*}{\makecell[c]{\textbf{Right wrist}}}
    & Omnicontrol & 0.631& 0.392& 10.836&  30.68& 6.65& 18.19\\ 
    & \textbf{Ours}     & \textbf{0.260}& \textbf{0.771}& \textbf{10.985}&  \textbf{0.57}& \textbf{0.03}& \textbf{0.57}\\ 
    
    \midrule
    \multirow{2}{*}{\makecell[c]{\textbf{Left foot}}}
    & Omnicontrol & 0.537& 0.389& \textbf{10.972}&  36.65& 7.88& 16.70\\ 
    & \textbf{Ours}   & \textbf{0.242}& \textbf{0.779}& 10.937&  \textbf{1.14}& \textbf{0.06}& \textbf{0.74}\\ 
    
    \midrule
    \multirow{2}{*}{\makecell[c]{\textbf{Right foot}}}
    & Omnicontrol & 0.575& 0.376& 10.893&  37.64& 7.74& 16.60\\ 
    & \textbf{Ours}     & \textbf{0.311}& \textbf{0.789}& \textbf{10.955}&  \textbf{1.42}& \textbf{0.07}& \textbf{0.82}\\

     \midrule
    \multirow{3}{*}{\makecell[c]{\textbf{Average}}}
    & Omnicontrol & 0.788& 0.379& 10.841&  14.33& 3.68& 8.54\\ 
    & TLControl & 0.487 & 0.751 & 10.716& \textbf{0.52}& 0.15 & 2.98\\
    & \textbf{Ours}     & \textbf{0.271}& \textbf{0.778}& \textbf{11.007}&  0.97& \textbf{0.05}& \textbf{0.64}\\

    \bottomrule
    \end{tabular}
    }
    \label{table:kit}
\end{table*}

\section{Experiments}

\subsection{Datasets and Evaluation Metrics}

\textit{Datasets.} We conduct comprehensive experiments on two mainstream text-to-motion dataset HumanML3D \cite{humanml3d} and KIT-ML \cite{plappert2016kit}. 
HumanML3D takes the motion sequences from the AMASS \cite{mahmood2019amass} and HumanAct12 \cite{humanact12} collections and re-annotates them. It contains 14,616 motion sequences annotated with 44,970 textual descriptions, comprising 5,371 distinct words. The total duration is 28.59 hours.
KIT-ML consists of 3,911 motion sequences and 6,353 sequence-level natural language descriptions. In our experiments, we leverage motion sequences with a maximum length of 196 frames.
HumanML3D is a large-scale dataset, while KIT-ML is relatively small. Thus, we can evaluate the performance of our methods in both resource-abundant and resource-limited settings.

\textit{Motion Fidelity Metrics.} For human motion, we aim to generate natural and diverse results harmoniously with the given textual description. Therefore, we adopt several evaluation metrics to validate the diversity and realism of generated motions, as proposed by \cite{humanml3d}. 

\begin{enumerate}
    \item \textit{Fréchet Inception Distance (FID)} measures the fidelity of generated motions. It measures the distance between the generated motion's and the real motion's distribution.
    \item \textit{R-precision} evaluates the matching degree between the generated motion and given textual description by calculating the Euclidean distance between the embeddings of motion and textual description.
    \item \textit{Diversity} measures the variance of the generated motions. It is calculated by choosing two subsets of 300 motion samples and averaging their Euclidean distances.
    \item \textit{Foot skating ratio} is a proxy for the physical plausibility of motions, following \cite{gmd, xie2023omnicontrol}. It measures the proportion of frames in which either foot skids more than a certain distance (2.5 cm) while maintaining contact with the ground (foot height $<$ 5 cm).
\end{enumerate}

\textit{Controllability Metrics.} We also adopt several controllability metrics, following the evaluation protocol from \cite{xie2023omnicontrol}. 
\begin{enumerate}
    \item \textit{Trajectory error} is the percentage of unsuccessful trajectories. The unsuccessful trajectory is defined as a trajectory where the location error of any keyframe exceeds a specific threshold.
    \item \textit{Location error} is the percentage of keyframe locations that are not reached within a threshold distance.
    \item \textit{Average error} measures the mean distance of locations between the generated motion and the given trajectories.
\end{enumerate}

\begin{figure*}[t]
    \vspace{0mm}
    \centering
       \includegraphics[width=1.0\linewidth]{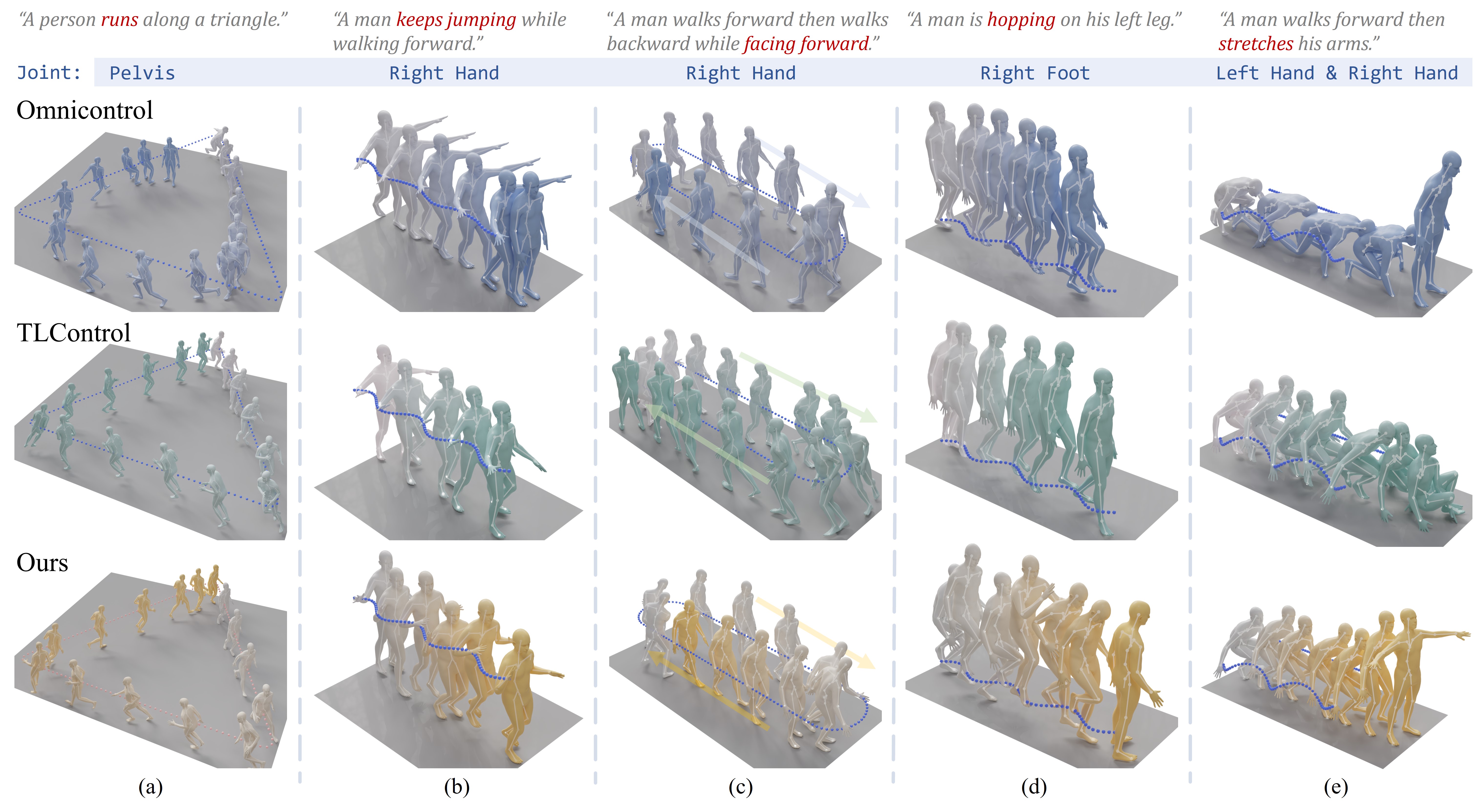}
       \vspace{-4mm}
        \caption{Qualitative comparisons between our method, Omnicontrol \cite{xie2023omnicontrol} and TLControl \cite{wan2023tlcontrol}. In order to compare their performances on out-of-distribution (OOD) trajectories, we adopt trajectories that are distinct from those in the current datasets.
            }
       \label{fig:compare}
    \vspace{0mm}
\end{figure*}

\subsection{Implementation Details} \label{subsec:implementation}

Here we present the detailed implementation of CMC. Both stages are built on a diffusion model, with a Transformer encoder architecture as their backbone \cite{mdm}. The Transformer encoder comprises 8 transformer layers with a feature dimension of 512. We use CLIP-ViT-B/32 as the text encoder in both stages. The resulting text embeddings are then concatenated at the start of the sequence of motion tokens. Both stages are trained for 400,000 iterations, with an initial learning rate of 2e-4 that decreases to 1e-5 at 200,000 iterations. The maximum diffusion step is set to 1000 for both stages.

\textit{Trajectory Control (Stage 1).} In this stage, we first normalize the pre-assigned trajectory and subsequently employ a trajectory encoder, specifically a three-layer Multi-Layer Perceptron (MLP), to map it into embedding space. The motion representation in this stage is the simplified representation that comprises pelvis angular velocity, pelvis linear velocity, and the local positions of non-pelvis joints. During training, we set six controllable joints: pelvis, left foot, right foot, head, left wrist, and right wrist . We then randomly select one to six of these joints for each training sample, ensuring flexibility and robustness in our model's learning. To accommodate varying lengths of controlled sequences, we dynamically choose the number of frames for each motion, ranging from one frame up to the maximum sequence length. 
To improve trajectory guidance, we utilize the L-BFGS optimizer. Given that the noise level in the motion data is relatively high during the initial stages of the denoising process, we implement a coarse-to-fine strategy to enhance the predicted motion. Specifically, during the first 990 steps, the optimization is conducted for 10 iterations with a learning rate of 0.5 to provide rough guidance for the denoising process. In the final 10 steps, the optimization is performed for 100 iterations with a learning rate of 0.1, allowing for more precise guidance.

\textit{Motion Completion (Stage 2).} This stage aims to complete the full-body motions with the pelvis and the control joints from the first stage as partial observations. During training, we use a spatial-temporal binary mask to mask the input motions by retaining the controlled joints as observations while leaving the other joints to remain noisy. To mitigate overfitting, SIM employs a randomized masking strategy: with 50\% probability, the input motions are masked, and with the remaining 50\% probability, they are replaced with pure noise without any known observations.

\nopagebreak[4]
\subsection{Trajectory-Controlled Human Motion Generation}
To evaluate the performance on trajectory-controlled human motion generation, we control the pelvis and five end-effectors (left foot, right foot, head, left wrist, and right wrist) following previous methods \cite{xie2023omnicontrol,wan2023tlcontrol} to ensure fair comparison. 
The main reason for selecting these six joints is as follows. The pelvis serves as the kinematic anchor of the human body and plays a pivotal role in determining the overall movement of a motion sequence. Meanwhile, the five end-effectors are the primary carriers of motion semantics and human-environment interaction: the hands perform manipulative actions (e.g., waving or touching objects), the feet govern locomotion (e.g., walking), and the head conveys intent (e.g., looking forward).
For specifying the trajectory conditions in real-world applications, users can either draw trajectories on a computer's interactive interface or sketch them in the air with VR controllers. To compare our CMC with state-of-the-art methods in the trajectory-controlled motion generation task, we evaluate it on the HumanML3D and KIT-ML datasets, using the ground-truth trajectories of joints as the condition.

We report our performance under a dense control setting (controlling 100\% of the frames). The quantitative comparison results for the HumanML3D and KIT datasets are presented in TABLE \ref{table:humanml3d} and TABLE \ref{table:kit}, respectively. Since MDM \cite{mdm}, PriorMDM \cite{priormdm}, and GMD \cite{gmd} can only control the pelvis, we report the performance while controlling the pelvis for fair comparisons. On the HumanML3D dataset, our model achieves state-of-the-art results when controlling the pelvis, as indicated by three key metrics: FID, R-precision, and average control error. For the single joint controlling except the pelvis, our method still achieves overall better results in FIDs. 
Among the current methods, those most closely related to our work are Omnicontrol \cite{xie2023omnicontrol} and InterControl \cite{intercontrol}. Both approaches utilize a ControlNet to integrate trajectory conditions into a basic motion diffusion model. 
When comparing with Omnicontrol and InterControl, we observe significant improvements in both the FID and R-precision metrics. Furthermore, there is a notable enhancement in controllability metrics, primarily due to our successful resolution of instability issues during trajectory guidance. 

We also report the metrics of controlling multiple joints in TABLE \ref{table:humanml3d} at the last row. Even when controlling multiple joints on all frames, our method shows a large improvement over Omnicontrol and is competitive with the state-of-the-art TLControl \cite{wan2023tlcontrol}.

On the KIT-ML dataset (TABLE \ref{table:kit}), we observe that our method consistently achieves state-of-the-art performance under the pelvis-control setting. Since TLControl didn't report the results on other non-pelvis joints, we only compare our method with Omnicontrol on the six main joints. As shown in TABLE \ref{table:kit}, we report a remarkable improvement over Omnicontrol on both motion quality and controllability metrics.

\vspace{1mm}
\textit{Qualitative Comparisons.}
In Fig. \ref{fig:compare}, we provide five groups of qualitative results to compare the performances on out-of-distribution (OOD) trajectories between our method, Omnicontrol \cite{xie2023omnicontrol}, and TLControl \cite{wan2023tlcontrol}. The given textual prompt and the controlled joint are presented on top of Fig. \ref{fig:compare}. As observed, our method consistently achieves better performance in terms of both motion quality and control accuracy. 
\begin{enumerate}
    \item In Fig. \ref{fig:compare} (a), which demonstrates the control under long distances, Omnicontrol fails to strictly follow the given triangle path; TLControl produces unrealistic motion (the person is floating in the air). 
    \item In Fig. \ref{fig:compare} (b), both Omnicontrol and TLControl fail to follow the instruction ``jumping". 
    \item In Fig. \ref{fig:compare} (c), our method correctly exhibits the action ``walk backward". 
    \item In Fig. \ref{fig:compare} (d), both Omnicontrol and TLControl generate a walking motion, with the character floating in mid-air, ignoring the ``hop" semantic. 
    \item In Fig. \ref{fig:compare} (e), Omnicontrol generates a human climbing on the ground, and TLControl fails to generate the standing and arm-stretching motion. Only our approach correctly exhibits the action ``stretch arms".
\end{enumerate}

Overall, our method demonstrates superior performance in text understanding and trajectory control. In contrast, Omnicontrol struggles significantly with trajectory control, while TLControl exhibits limited text understanding with out-of-distribution text data.

\begin{figure}[t]
    \vspace{-2mm}
  \centering
    \includegraphics[width=\linewidth]{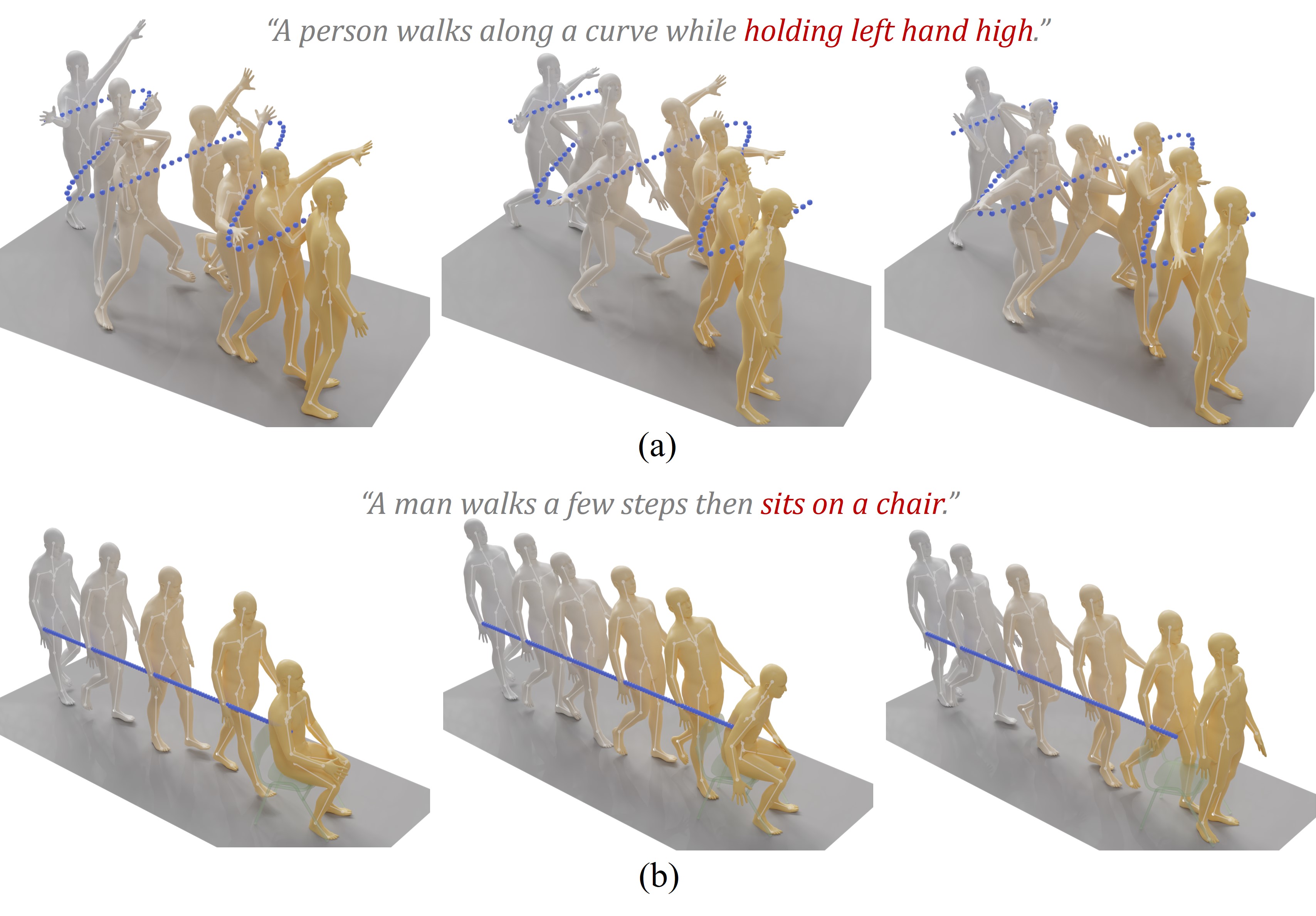}
    
    \caption{Two visual comparisons (a) and (b) to qualitatively prove the existence of the conflict between text and trajectory conditions. The difference between the three columns is the number of joints that passed to the second stage, including: only the controlled joints and the pelvis (left), the controlled joints and joints belonging to the torso (middle), and all joints (right).}
    \vspace{-2mm}
  \label{fig:ablation}
\end{figure}

\begin{table}[t] 
    \vspace{0mm}
    \centering
    \caption{Ablation results of different combinations of proposed components on the HumanML3D testing set. Each result is the average of the performance metrics for the pelvis and five end-effectors.}
    \large
    \resizebox{0.48\textwidth}{!}{
    \begin{tabular}{c|cccc|ccc}
    \toprule
        \multicolumn{1}{p{2.0cm}|}{\centering Item} 
        & \multicolumn{1}{p{1.0cm}}{\centering Two \\ stage} 
        & Optimizer 
        & \multirow{1}{*}{SIM} 
        & \multicolumn{1}{p{2.3cm}|}{\centering Joints Passed \\ to the 2nd stage}
        & FID $\downarrow$ 
        & \multicolumn{1}{p{2.0cm}}{\centering R-precision  \\ (Top-3) $\uparrow$}
        & \multicolumn{1}{p{1.7cm}}{\centering Avg. Err. $\downarrow$  (cm)}  \\

    \midrule
        \circled{1} & \XSolidBrush &  SGD &  - & - & 0.385 & 0.677 & 5.13\\
        \circled{2} & \XSolidBrush & L-BFGS &  - & - & 0.364 & 0.698 & 1.88\\
        \circled{3} & \Checkmark &  SGD &  \Checkmark & controlled joints & 0.165 & 0.759& 1.63\\
    \midrule
        \circled{4} & \Checkmark & L-BFGS & \XSolidBrush & controlled joints &  0.176 &  0.749& 0.61 \\
        \circled{5} & \Checkmark & L-BFGS & \Checkmark & all joints& 0.183 & 0.715& 0.61\\
        \circled{6} & \Checkmark & L-BFGS & \Checkmark & torso joints&  0.139 &  0.740& 0.61\\
    \midrule
        \circled{7} & \Checkmark & L-BFGS & \Checkmark & controlled joints & \textbf{0.119} & \textbf{0.775}& \textbf{0.61} \\

    \bottomrule
    \end{tabular}
    }
    \vspace{-1mm}

    \label{table:ablation}
\end{table}

\subsection{Ablation Study on Each Component} \label{subsec:ablation_component}

We perform several ablation experiments, as shown in TABLE \ref{table:ablation}, to justify the effectiveness of each component in our framework. From these, we can infer several key conclusions:
\begin{enumerate}
    \item \textit{Tackling the key conflicts with our two-stage framework evidently improves the overall performance.}
    The results from experiments \circled{1} and \circled{3} clearly demonstrate that~our decoupled framework significantly enhances motion quality and control accuracy, even without using a superior optimizer. Omnicontrol shares the same settings as experiment \circled{1}. While Omnicontrol outperforms experiment \circled{1} in terms of FID and R-precision, this improvement can be attributed to the use of ControlNet. In contrast, our framework utilizes a common MDM and achieves noteworthy results even using the same SGD optimizer (\circled{3}).
    \item \textit{SIM notably improves the generalization ability of the motion diffusion model.}
    The comparison from experiments \circled{4} and \circled{7} indicates that our proposed SIM indeed improves the generalization ability of the network in the second stage, especially reflected in the two metrics FID and R-precision.
    \item \textit{Naively applying a superior optimizer cannot lead to a better comprehensive performance.}
    From the experiments \circled{1} and \circled{2}, we conclude that simply using a better optimizer does not significantly improve overall performance. Although the L-BFGS optimizer demonstrates enhanced accuracy in controlling motion trajectories, it does not substantially affect the quality and realism of generated motions as measured by FID and R-precision. This suggests that although optimization techniques are important, they alone may not be sufficient to address the challenges in text and trajectory-controlled human motion generation.
    \item \textit{The conflict between the text and trajectory conditions affects the realism of generated motion.}
    Note that the output of the first stage includes the local positions of all joints. To avoid conflict between text and trajectory conditions to the greatest extent, we only pass the trajectory of the pelvis and controlled joints into the second stage. 
    To validate this operation, we conduct experiments \circled{5}, \circled{6}, and \circled{7}. The only difference among these experiments is the number of joints that are passed to the second stage as partial observations. We can observe that as the number of joints increases, the FID and R-precision gradually decline. This quantitatively confirms that this conflict affects the quality of the generated motion. Moreover, we also qualitatively demonstrate the existence of such conflict via visualization examples in the following paragraph.
\end{enumerate}

We also provide qualitative comparisons in Fig. \ref{fig:ablation}. 
In Fig. \ref{fig:ablation} (a), as the number of joints sent to the second stage (from left to right) increases, the generated motion gradually deviates from the intended semantics of the text description. When all the body joints from the first stage are passed to the second stage, the person continues walking without raising their left hand in any of the frames.
In Fig. \ref{fig:ablation} (b), the torso from the first stage appears unrealistic, as the person's upper body leans back while walking. Moreover, when all joints from the first stage are sent to the second stage, the person fails to execute the ``sit" action correctly. These results indicate that, due to the conflict between the text and trajectory conditions, the network may experience difficulties in fully understanding the text prompt, resulting in unnatural motions. Thus, we only pass the controlled joints' representation to the second stage and let the second stage conduct motion completion.

\begin{table*}[t!]
    \centering
    \vspace{-2mm}
    \caption{Quantitative results under the text-to-motion generation setting on the HumanML3D testing set. \textbf{Bold} indicates the best result, while \underline{underscore} refers to the second-best. ``$\rightarrow$" denotes closer to that of the real data is better.}
    \resizebox{0.90\textwidth}{!}{
    \begin{tabular}{cccccccc} 
    \toprule
    \multirow{2}{*}{Datasets} & \multirow{2}{*}{Methods}  & \multicolumn{3}{c}{R-Precision$\uparrow$} & \multirow{2}{*}{FID$\downarrow$} & \multirow{2}{*}{Diversity$\rightarrow$} & \multirow{2}{*}{MM-Dist$\downarrow$}\\
    \cline{3-5}
       ~& ~ & Top 1 & Top 2 & Top 3 \\

    \midrule
    \multirow{5}{*}{\makecell[c]{Human\\ML3D}}  
        & \textbf{Real Motion}   & \et{0.511}{.003}  & \et{0.703}{.003} & \et{0.797}{.002} & \et{0.002}{.000} & \et{9.503}{.065} & \et{2.974}{.008}\\ 
        \cline{2-8}
        & MDM~\cite{mdm}~\pub{ICLR23}           & \et{0.418}{.005}  & \et{0.604}{.005} & \et{0.707}{.004} & \et{0.489}{.025} & \etb{9.559}{.086} & \et{3.630}{.023}\\ 
        & MotionDiffuse~\cite{zhang2022motiondiffuse}~\pub{TPAMI24} & \etb{0.491}{.001}  & \etb{0.681}{.001} & \etb{0.782}{.001} & \et{0.630}{.001} & \ets{9.410}{.049} & \etb{3.113}{.001}\\ 
        & MLD~\cite{mld}~\pub{CVPR23}           & \ets{0.481}{.003}  & \ets{0.673}{.003} & \ets{0.772}{.002} & \ets{0.473}{.027} & \et{9.724}{.082} & \ets{3.196}{.010}\\ 
        \cline{2-8} 
        
        & MDM (ours) &   \et{0.476}{.002} & \et{0.668}{.002} & \et{0.767}{.002} & \etb{0.261}{.069} & \et{9.894}{..091} & \et{3.228}{.012}\\ 

    \midrule
    \multirow{5}{*}{\makecell[c]{KIT-\\ML}}   
        & \textbf{Real Motion}   & \et{0.424}{.005}  & \et{0.649}{.006} & \et{0.779}{.006} & \et{0.031}{.004} & \et{11.08}{.097} & \et{2.788}{.012}\\ 
        \cline{2-8}
        & MDM~\cite{mdm}~\pub{ICLR23}           & \et{0.403}{.005}  & \et{0.606}{.004} & \et{0.731}{.004} & \et{0.513}{.045} & \et{10.85}{.109} & \et{3.096}{.023}\\ 
        & MotionDiffuse~\cite{zhang2022motiondiffuse}~\pub{TPAMI24}   & \ets{0.417}{.004}  & \ets{0.621}{.004} & \ets{0.739}{.004} & \et{1.954}{.064} & \etb{11.10}{.143} & \etb{2.958}{.005}\\ 
        & MLD~\cite{mld}~\pub{CVPR23}            & \et{0.390}{.008}  & \et{0.609}{.008} & \et{0.734}{.007} & \etb{0.404}{.027} & \et{10.80}{.117} & \et{3.204}{.027}\\ 
        \cline{2-8}
        & MDM (ours) &   \etb{0.429}{.006}  & \etb{0.637}{.005} & \etb{0.751}{.004} & \ets{0.439}{.022} & \ets{10.98}{.068} & \ets{3.033}{.016}\\ 
    
    \bottomrule
    \end{tabular}
    }
    \vspace{-2mm}

    \label{table:sim_on_t2m}
\end{table*}

\begin{figure}[t]
  \vspace{-7mm}
  \centering
  \subfloat[with SIM]{
  \includegraphics[width=0.48\linewidth]{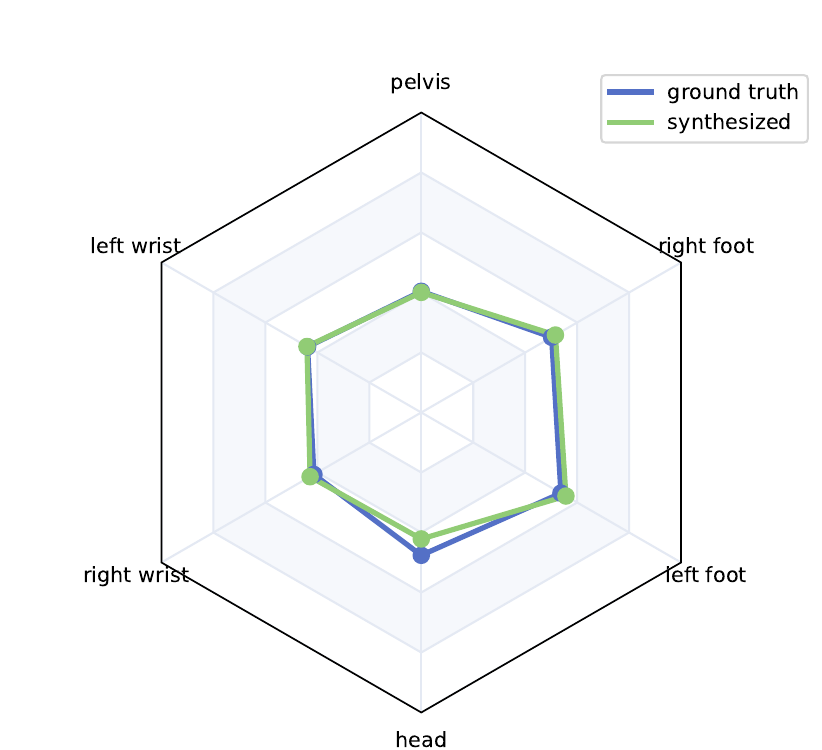}%
  }
  \hfill
  \subfloat[without SIM]{
  \includegraphics[width=0.48\linewidth]{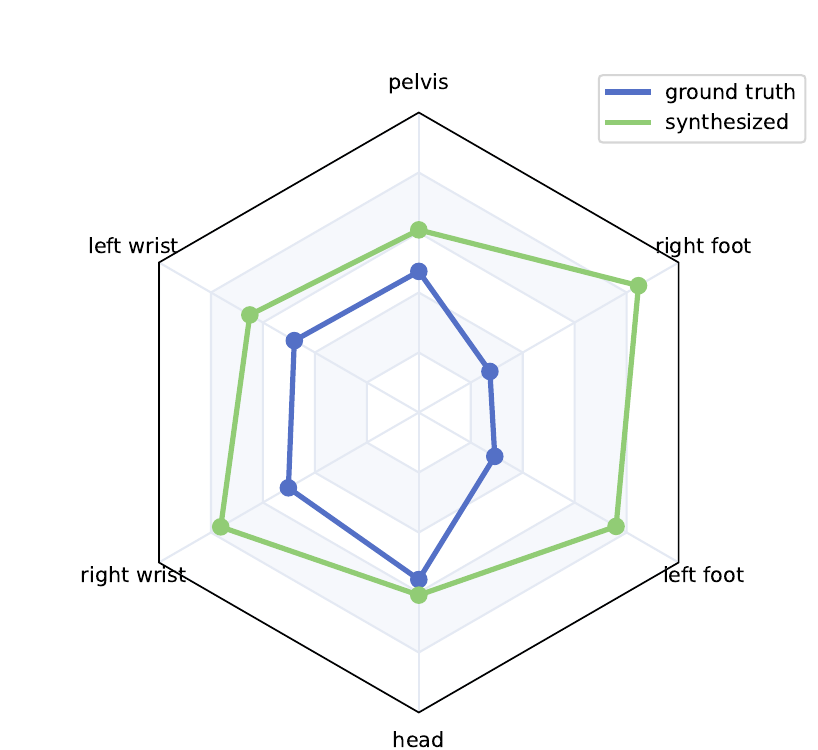}%
  }
  
  \caption{Comparisons on FID scores with and without SIM. Each figure plots the FIDs when controlling the pelvis and five end-effectors individually. The model with SIM consistently performs well based on both ground truth and synthesized partial observations.}
  \label{fig:sim}
  \vspace{-4mm}
\end{figure}

\subsection{In-Depth Analysis of 
SIM} \label{sec:sim}

\subsubsection{FID of Six Joints with and without SIM} 

To more effectively illustrate the contribution of our proposed SIM on various controlled joints, we trained the diffusion inpainting model with and without SIM, respectively. We then evaluated the two models using the same first-stage model on the HumanML3D testing set. Fig. \ref{fig:sim} shows the FID scores for the pelvis and five end-effectors using a hexagon map. Each figure displays the FID scores conditioned on two types of partial observations: those from the test set ground truth (in blue) and those synthesized by the first stage (in green). The comparison between Fig. \ref{fig:sim} (a) and (b) reveals that the model with SIM yields negligible differences in the FID scores between the ground truth and the synthesized partial observations. Conversely, the performance gap is much more pronounced for the model without SIM. This indicates that SIM effectively enhances the diffusion model's generalization ability to out-of-domain conditions.

\subsubsection{Boosting Text-to-Motion Generation Performance via Plug-and-Play SIM} 

Through the ablation experiments in Section \ref{subsec:ablation_component}, we demonstrate that our proposed SIM can alleviate the degree of overfitting of the model to the known partial observations in the motion inpainting task and make the model well generalized to unseen text and trajectory data. Thus, SIM can lead to improved FID and R-precision.
Here, we further demonstrate that SIM can boost the generalization ability of diffusion models. SIM trains the motion diffusion model in a multi-task learning manner. With multi-task training, the motion diffusion model can capture common representations between the motion inpainting task and the text-to-motion generation task, thereby improving the model's generalization across these two tasks.

To prove the viewpoint above, we take our MDM trained with SIM ( denoted as our MDM for brevity ) as an example and evaluate it under the text-to-motion generation setting.  
Since SIM is tailored for diffusion models, we compare our MDM to the state-of-the-art diffusion-based backbones under the text-to-motion generation setting. 
In TABLE \ref{table:sim_on_t2m}, our MDM achieves significantly better results than the original MDM in terms of FID and R-precision on both the HumanML3D and KIT-ML datasets. 
On the HumanML3D dataset, our MDM achieves the best FID scores and a higher degree of text-to-motion matching (R precision and MM-Dist). 
On the KIT-ML dataset, our MDM achieves the best R precision scores and competitive results in FID, diversity, and MM-Dist, which demonstrate that SIM's effectiveness is more significant in a resource-limited setting.

\begin{figure}[t]
  \vspace{-4mm}
  \centering

  \subfloat[sample 1]{
  \includegraphics[width=0.48\linewidth]{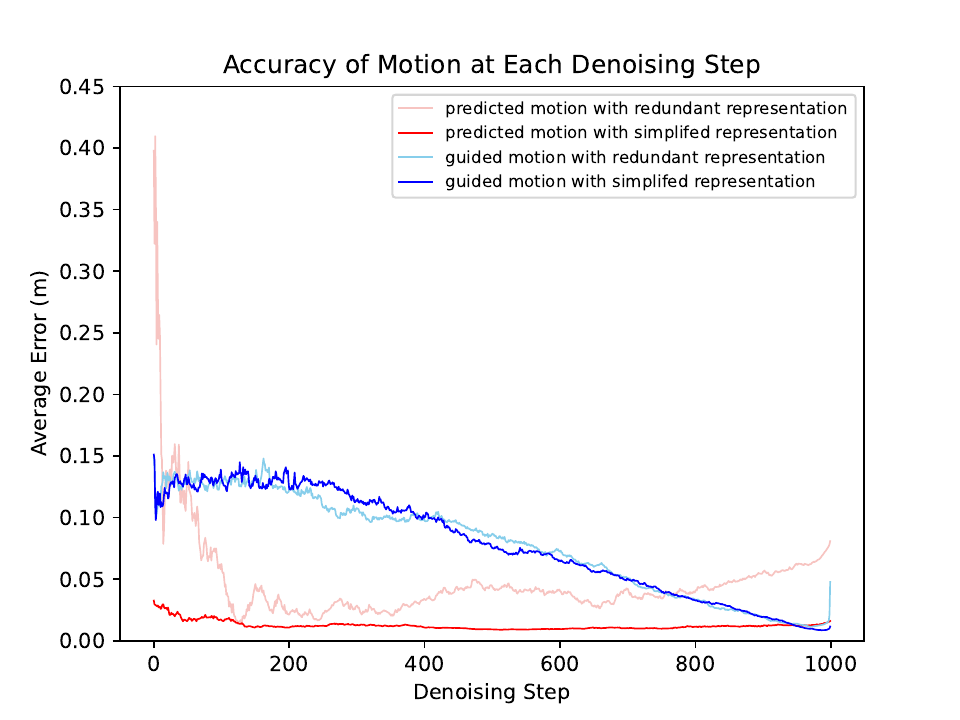}%
  }
  \hfill
  \subfloat[sample 2]{
  \includegraphics[width=0.48\linewidth]{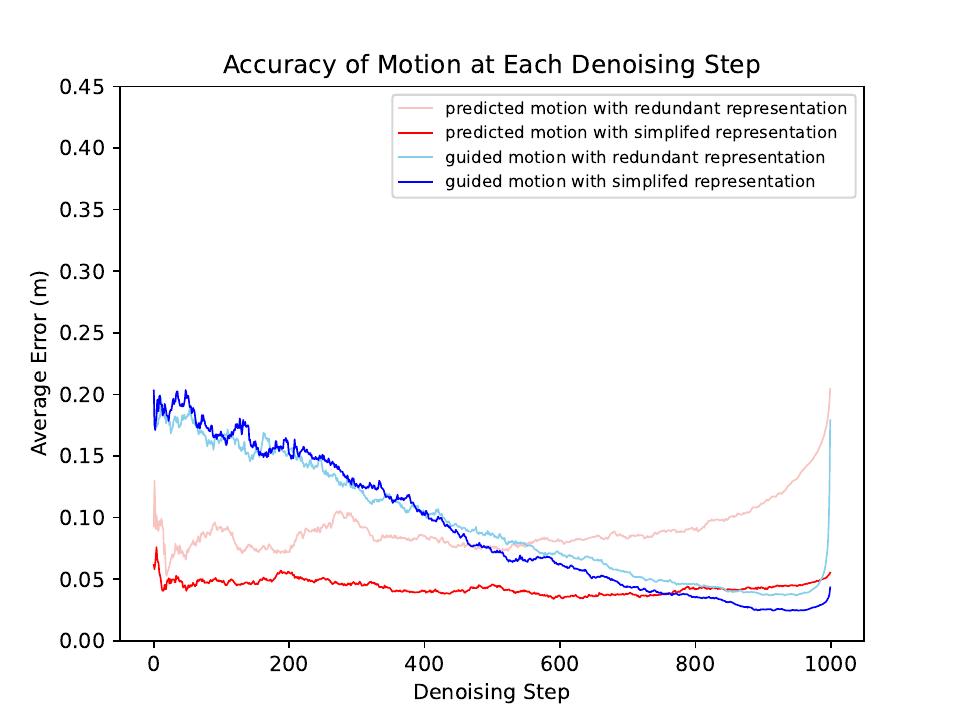}%
  }
  
  \caption{Average control error across denoising steps. Darker-colored and lighter-colored curves indicate the use of simplified and redundant representations, respectively.}
  \vspace{-6mm}
  \label{fig:1000} 
  
\end{figure}

\begin{figure}[t]
  \vspace{0mm}
  \centering
  \subfloat[Predicted Motion]{
  \includegraphics[width=0.48\linewidth]{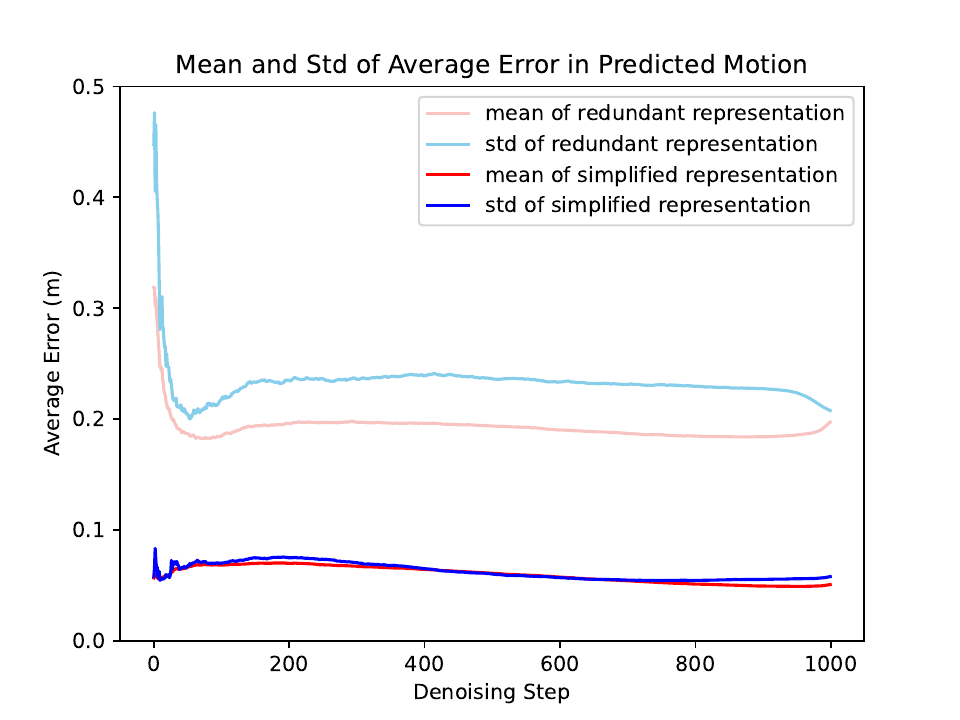}%
  }
  \hfill
  \subfloat[Guided Motion]{
  \includegraphics[width=0.48\linewidth]{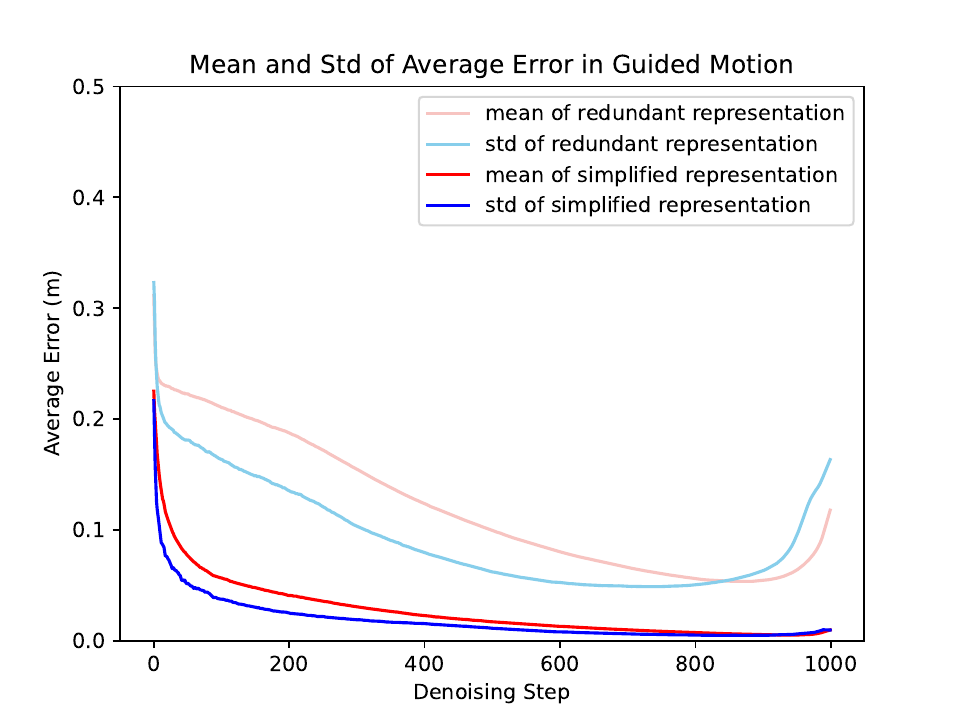}%
  }
  
  \caption{Statistical mean and standard deviation of the control error across denoising steps. Darker-colored and lighter-colored curves indicate the use of simplified and redundant representations, respectively.}
  \label{fig:statistic}
  \vspace{0mm}
\end{figure}

\subsection{In-Depth Analysis of Redundant Representations} \label{sec:redundant}

\subsubsection{Visualization of Instability in Redundant Representation}
We present the curves of control error for each step of the denoising process of two samples, (a) and (b), shown in Fig. \ref{fig:1000}. 
The dark curves indicate the simplified representation, while the light curves indicate the redundant representation. The red curves represent predicted motion by the network, and the blue curves reflect the guided motion by trajectory guidance. 
As the generation progresses, the error of the predicted motion (light red) from the model with redundant representation becomes noticeably unstable, generally showing an upward oscillation trend. Toward the end of the denoising process, the inconsistency in redundant representation becomes apparent, leading to a rise in control error.  
In contrast, the predicted motion's control error (dark red) of the model with simplified representation remains stable within a certain range. Benefiting from the stability of predicted motion, the final accuracy of the model with simplified representation can achieve better results than that of the model with redundant representation.

\subsubsection{Statistical Analysis of Different Representations}

To provide a more statistically robust demonstration of the instability issues of redundant representations, we plot the mean and standard deviation (std) of control errors (including errors of both predicted motion and guided motion) across 1000 denoising steps. These statistical metrics are computed using all samples from the test set. To ensure a fair comparison and eliminate performance gains from L-BFGS optimizer, we use the SGD optimizer and fix the number of guidance iterations to 10 at each denoising step following \cite{xie2023omnicontrol}.
As shown in Fig. \ref{fig:statistic} (a), the mean control error curves of the simplified representation are significantly lower than those of the redundant representation across all denoising steps. Meanwhile, the standard deviation of the simplified representation remains substantially smaller than that of the redundant representation. Lower mean values indicate better trajectory following, and smaller standard deviations reflect a more consistent and stable trajectory following. These results confirm that our proposed simplified representation effectively mitigates the instability inherent to redundant representations. 
We further validate the conflict between text and trajectory conditions through Fig. \ref{fig:statistic} (b). For the redundant representation, the control error of the guided motion exhibits a clear trend of first decreasing and then increasing as denoising progresses. In the early stages of denoising, the generated motion is low-quality and disordered, so trajectory guidance plays a dominant role in reducing errors. Toward the end of denoising, the text condition gradually takes precedence: the generated motion tends to align more closely with textual semantics, which deviates from the pre-defined trajectory and leads to a rise in control error. This phenomenon directly validates our analysis of the conflict between text and trajectory conditions. In contrast, the mean and standard deviation curves exhibit a trend of smoothly decreasing when using the simplified representation, finally achieving precise control.

\subsubsection{Control Performance of whether Using Text Condition}
\begin{figure}[t]
    \vspace{-2mm}
  \centering
    \includegraphics[width=\linewidth]{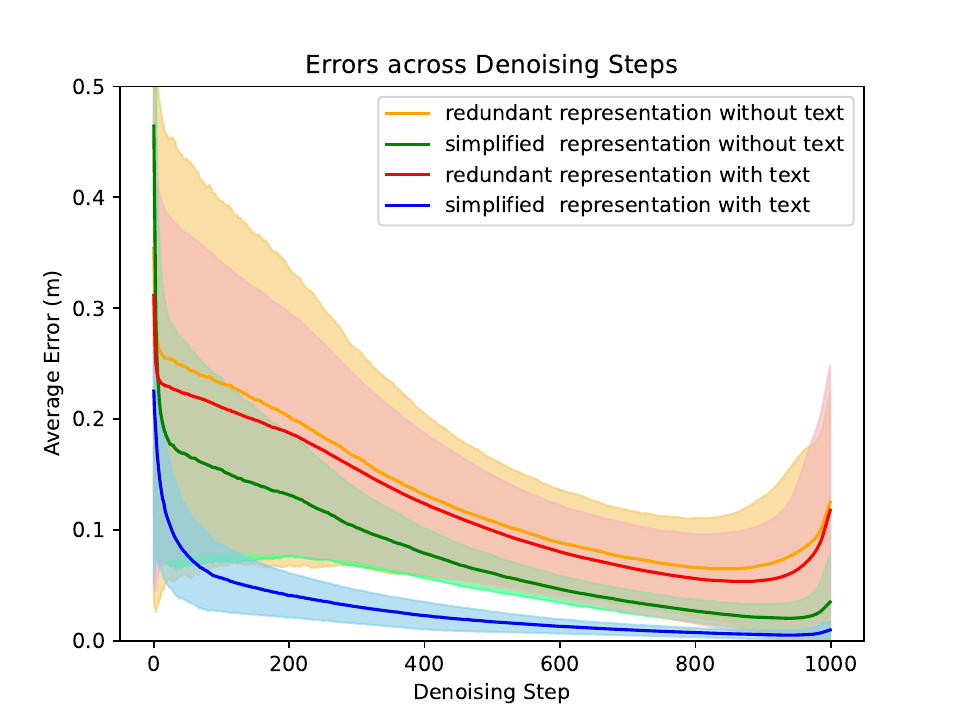}
    
    \caption{Errors across all denoising steps.The solid lines denote the mean value, and the areas with light color denote the band of standard deviation, illustrating the fluctuation centered around the mean value.}
    \vspace{0mm}
    \label{fig:error_4_color}
\end{figure}

We present a mean plot with a standard deviation band in Fig. \ref{fig:error_4_color} to illustrate the control error across denoising steps. 
In Fig. \ref{fig:error_4_color}, the solid lines represent the mean values, while the light-colored areas indicate the standard deviation band, highlighting the fluctuations around the mean.
It is observed that the result of the redundant representation without conditioning on text exhibits the largest instability, which validates the analysis of the inconsistency inherent in the redundant representation. 
In contrast, when text is included as a condition, the control error improves slightly and becomes more stable. We believe this is because the distribution of unconditional motion generation is much larger and more indeterminate than that of text-to-motion generation during generation. Additionally, the texts in the datasets align closely with the trajectory, resulting in a slight performance improvement. Conversely, if the text deviates significantly from the trajectory, we anticipate a decline in control performance. 
Regarding the simplified representation, it consistently outperforms the redundant representation and demonstrates greater stability across all denoising steps, highlighting its effectiveness.

\subsubsection{Ablation Experiments on the Components in Redundant Representation} In TABLE \ref{table:ablation_redundant}, we provide the control accuracy of three variants of using different components in redundant representation. As the components used increase, the control accuracy degrades gradually.

\begin{table}[t]
    \centering
    \caption{ Average errors on 1 frame- and 100\% frame-control settings. Experiments are conducted on the HumanML3D testing set.}
    \resizebox{0.45\textwidth}{!}{
    \begin{tabular}{c|ccccc} 
    \toprule

    Joint & position & rotation & velocity & 1 frame (cm) & 100\% frames (cm) \\
    \hline
    \multirow{3}{*}{\makecell[c]{Average}}  
    & \Checkmark &            &            & \textbf{0.50} & \textbf{1.63} \\
    & \Checkmark & \Checkmark &            & 0.62 & 3.61 \\
    & \Checkmark & \Checkmark & \Checkmark & 0.69 & 5.13 \\
    
    \bottomrule
    \end{tabular}
    }

    \vspace{-2mm}
    \label{table:ablation_redundant}
\end{table}

\begin{table}[h]  
    \large
    \centering
    \caption{
        Evaluation of our CMC with DDIM sampling on the HumanML3D testing set.
    }
    \resizebox{0.48\textwidth}{!}{
    \begin{tabular}{c|c|cc|cccc} 
    \toprule
    
    Joint
    & Method
    & S1 steps 
    & S2 steps 
    & FID $\downarrow$ 
    & \multicolumn{1}{p{2.0cm}}{\centering R-precision \\ (Top-3) $\uparrow$} 
    & Diversity $\rightarrow$ 
    & \multicolumn{1}{p{1.6cm}}{\centering Avg. err. $\downarrow$  (cm)} \\
    
    \midrule
    
    Real & - & - & - & 0.002  & 0.797 & 9.503  & 0.00   \\ 
    
    \midrule
    \multirow{6}{*}{\makecell[c]{Pelvis}} 
    & TLControl & - & -  & 0.271 & 0.779 & 9.569 & 1.08 \\
    \cline{2-8}
    & \multirow{5}{*}{\makecell[c]{CMC}} 
    & 50 & 50     & 0.176 & \tb{0.789} & 9.773  & \tb{0.49}  \\  
    & & 50 & 100    & 0.146 & 0.788 & 9.774  & 0.49  \\  
    & & 100 & 50   & 0.175 & 0.788 & 9.761  & 0.50 \\   
    & & 100 & 100  & 0.160 & 0.789 & 9.634  & 0.51 \\ 
    & & 1000 & 1000  & \tb{0.097} & 0.784 & \tb{9.539} & 0.51  \\

    \midrule
    \multirow{6}{*}{\makecell[c]{Left \\ foot}} 
    & TLControl & - & -  & 0.368 & 0.768 & 9.774 & 1.14 \\
    \cline{2-8}
    & \multirow{5}{*}{\makecell[c]{CMC}} 
    & 50 & 50   & 0.198 & \tb{0.787} & 9.895  & 0.84  \\  
    & & 50 & 100    & 0.176 & 0.779 & 9.858  & \tb{0.83}  \\  
    & & 100 & 50   & 0.184 & 0.783 & 9.835  & 0.84 \\   
    & & 100 & 100  & 0.181 & 0.777 & 9.747  & 0.85 \\ 
    & & 1000 & 1000  & \tb{0.100} & 0.771 & \tb{9.598} & 0.85  \\

    \midrule
    \multirow{5}{*}{\makecell[c]{Right \\ foot}} 
    & TLControl & - & - & 0.361 & 0.775 & 9.778 & 1.16 \\
    \cline{2-8}
    & \multirow{5}{*}{\makecell[c]{CMC}} 
    & 50 & 50   & 0.220 & 0.786 & 9.770  & \tb{0.84}  \\  
    & & 50 & 100    & 0.180 & 0.785 & 9.788  & 0.85  \\  
    & & 100 & 50   & 0.204 & \tb{0.790} & 9.801  & 0.87 \\   
    & & 100 & 100  & 0.180 & 0.783 & 9.658  & 0.86 \\ 
    & & 1000 & 1000  & \tb{0.107} & 0.774 & \tb{9.600}  & 0.87 \\

    \midrule
    \multirow{5}{*}{\makecell[c]{Head}} 
    & TLControl & - & - & 0.279 & 0.778 & 9.606 & 1.10 \\
    \cline{2-8}
    & \multirow{5}{*}{\makecell[c]{CMC}} 
    & 50 & 50   & 0.197 & \tb{0.788} & 9.769  & 0.74  \\  
    & & 50 & 100    & 0.175 & 0.784 & 9.784  & \tb{0.72}  \\  
    & & 100 & 50   & 0.218 & 0.781 & 9.779  & 0.73 \\   
    & & 100 & 100  & 0.204 & 0.786 & 9.717  & 0.74 \\ 
    & & 1000 & 1000  & \tb{0.105} & 0.780 & \tb{9.567}  & 0.75 \\

    \midrule
    \multirow{5}{*}{\makecell[c]{Left \\ wrist}} 
    & TLControl & - & -  & \tb{0.135} & 0.789 & 9.757 & 1.08 \\
    \cline{2-8}
    & \multirow{5}{*}{\makecell[c]{CMC}} 
    & 50 & 50   & 0.225 & \tb{0.794} & 9.675  & 0.84  \\  
    & & 50 & 100    & 0.215 & 0.785 & 9.550  & \tb{0.83}  \\  
    & & 100 & 50   & 0.220 & 0.787 & \tb{9.488}  & 0.87 \\   
    & & 100 & 100  & 0.228 &0.788 & 9.555  & 0.86 \\ 
    & & 1000 & 1000  & 0.159 & 0.763 & 9.453 & 0.87 \\

    \midrule
    \multirow{5}{*}{\makecell[c]{Right \\ wrist}} 
    & TLControl & - & -  & \tb{0.137} & 0.787 & 9.734 & 1.09 \\
    \cline{2-8}
    & \multirow{5}{*}{\makecell[c]{CMC}} 
    & 50 & 50   & 0.195 & 0.787 & 9.714  & 0.84  \\  
    & & 50 & 100    & 0.200 & \tb{0.794} & 9.757  & \tb{0.82}  \\  
    & & 100 & 50   & 0.208 & 0.787 & 9.689  & 0.83 \\   
    & & 100 & 100  & 0.209 & 0.787 & 9.674  & 0.84 \\ 
    & & 1000 & 1000  & 0.149 & 0.775 & \tb{9.627}  & 0.85 \\

    \bottomrule
    \end{tabular}
    }
    \vspace{-2mm}

    \label{table:ddim}
\end{table}

\begin{table}[t]
    \centering
    \large
    \vspace{0mm}
    
    \caption{Performance Benchmarks Comparison with Previous Works. This table include results on motion quality and running efficiency.}
    \resizebox{0.45\textwidth}{!}{
    \begin{tabular}{c|cccccc}
    \toprule
    Methods
    & FID $\downarrow$
    & \multicolumn{1}{p{2.1cm}}{\centering R-precision \\ (Top-3) $\uparrow$}
    & \multicolumn{1}{p{2.1cm}}{\centering Avg. err. $\downarrow$  (cm)}
    & \multicolumn{1}{p{2.9cm}}{\centering Latency Under \\ Batch 32 (s) $\downarrow$}
    & \multicolumn{1}{p{2.1cm}}{\centering Average \\ Latency $\downarrow$}
    & FPS $\uparrow$ \\

    \midrule
    MDM & 0.698 & 0.602 & 59.59 & 33 & 1.03 & 190\\
    GMD & 0.576& 0.665 & 14.39 & 96 & 3.00 & 65\\
    Omnicontrol & 0.292 & 0.692 & 3.97 & 105 & 3.28 & 60 \\
    TLControl & 0.258 & \underline{0.779} & 1.10 & \textbf{9} & \textbf{0.28} & \textbf{700} \\
    CMC & \textbf{0.119} & 0.775 & \textbf{0.61} & 90 & 2.81 & 70 \\
    CMC-DDIM & \underline{0.193} & \textbf{0.785} & \underline{0.78} & \underline{13} & \underline{0.41} & \underline{478} \\

    \bottomrule
    \end{tabular}
    }
    \vspace{-1mm}
    
    \label{tab:infer_time}
\end{table}

\subsection{Accelerating CMC with and Real-Time Performance}

In this section, we evaluate CMC using DDIM sampling. As shown in Table \ref{table:ddim}, we assess CMC under four combinations of the total denoising steps in both stages. In Table \ref{table:ddim}, it is observed that the FID scores of CMC with DDIM sampling are relatively lower than those achieved with DDPM sampling. Additionally, as the total number of sampling steps decreases, there is a slight improvement in R precision but an overall decline in diversity. When comparing CMC with TLControl, despite having worse FID scores and diversities than DDPM sampling, CMC with DDIM sampling still outperforms TLControl overall.
We also provide the real-time performance benchmark comparisions with previous works in TABLE \ref{tab:infer_time}, including latency under a batch size of 32, average latency, and FPS, which demonstrate a high practicability of CMC.

\begin{figure}[t!]
    \vspace{0mm}
  \centering
    \includegraphics[width=0.9\linewidth]{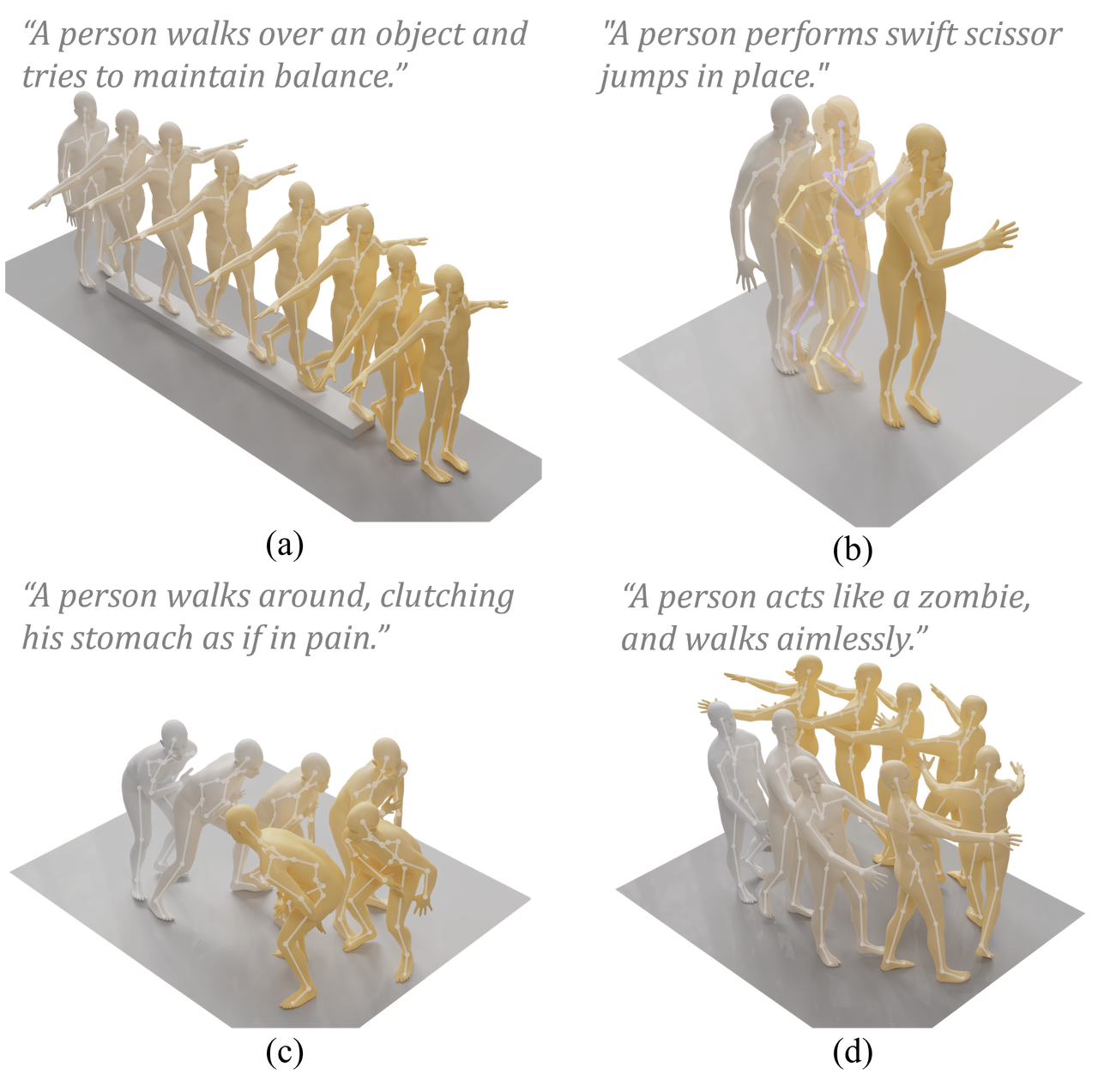}
    \vspace{-3mm}
    \caption{Qualitative visualizations of motions conditioned on text only.}
    \vspace{-3mm}

  \label{fig:supp_text_only}
\end{figure}

\begin{figure}[t!]
    \vspace{-4mm}
  \centering
    \includegraphics[width=\linewidth]{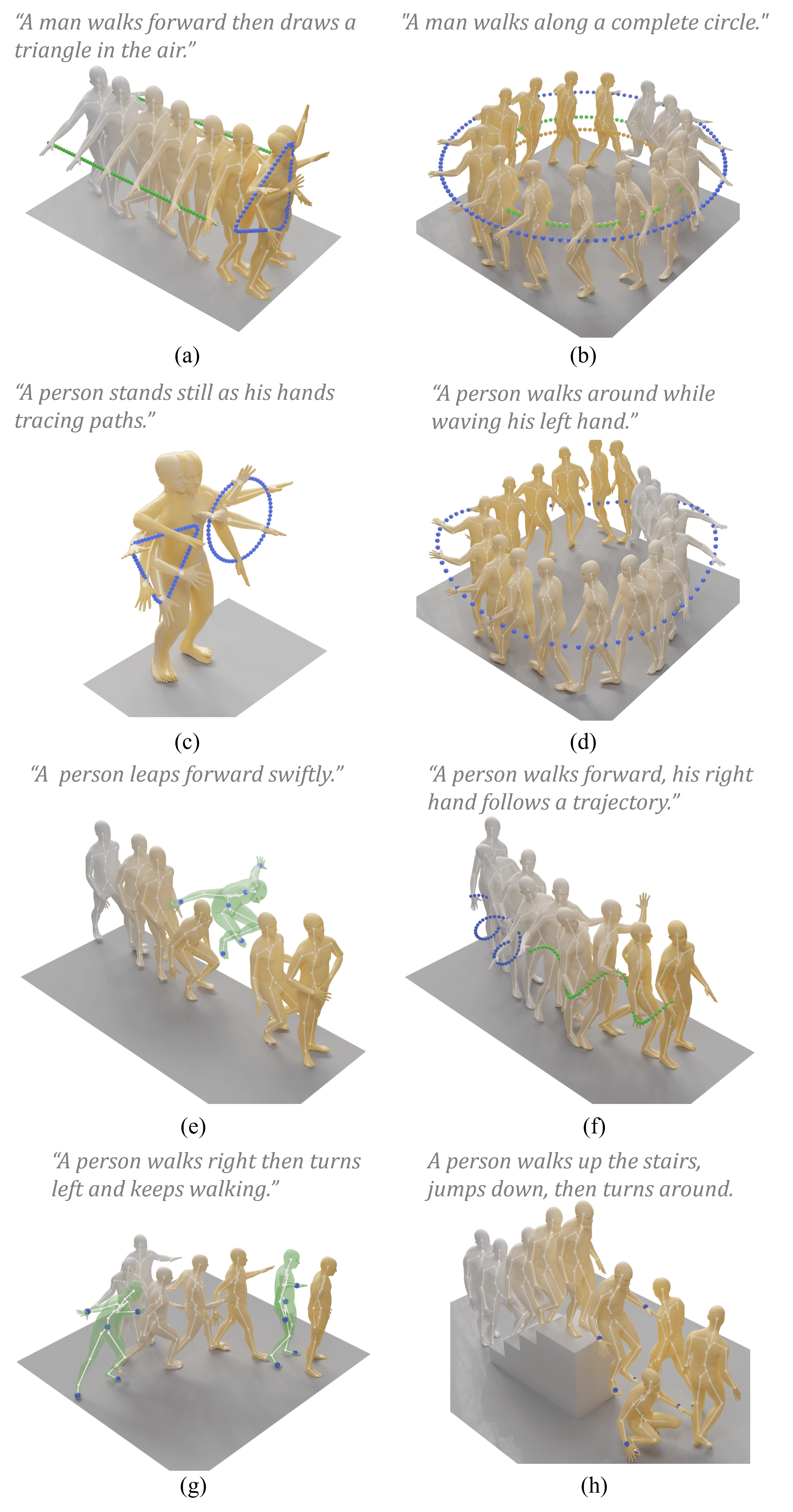}
    \vspace{-5mm}
    \caption{Qualitative visualizations of motions conditioned on both text and trajectory.}
    \vspace{-5mm}
  \label{fig:supp_text_traj}
\end{figure}

\vspace{-4mm}
\subsection{More Qualitative Visualization Examples}

We provide more visualization examples to validate the effectiveness of our method and the quality of the generated motions.  
We present visualization examples conditioned on text only in Fig. \ref{fig:supp_text_only} and those conditioned on both text and trajectory in Fig. \ref{fig:supp_text_traj}. To improve visual clarity, we provide supporting video demonstrations for all visualizations in this work.

\section{Discussions on Motion Realism and Control Fidelity}

We emphasize that the trade-off between motion realism and control accuracy exists in previous works Omnicontrol~\cite{xie2023omnicontrol} and TLControl~\cite{wan2023tlcontrol}. Specifically, this phenomenon is more easily observed in the motions that involve continuous moving, such as ``running quickly", because controlling this type of motion requires precise coordination across body parts. Enforcing strict trajectory adherence without considering global motion coherence can lead to artifacts such as foot skating or body floating.

As illustrated in Fig.~\ref{fig:compare}, Omnicontrol adopts a control-while-generate paradigm, resulting in insufficient trajectory following and weak semantic alignment caused by the conflict between text and trajectory condition. TLControl follows a generate-then-control approach, where the post-hoc control adjustments may distort the original motion structure, resulting in an unnatural motion. 
To better illustrate this, we use Fig.~\ref{fig:compare} (a) as a representative example: a large triangular trajectory (4 meters per side) is specified for the pelvis. Omnicontrol fails to follow the path tightly, while TLControl achieves high trajectory accuracy but produces unnatural results: the person's pelvis closely follows the trajectory but floats in mid-air, indicating that strong guidance lacking semantic understanding can lead to physically implausible motion patterns.
In contrast, our CMC adopts a control-then-generate paradigm. We first generate and fix the controlled joints, then complete the remaining joints using a diffusion-based inpainting model. This decoupling allows CMC to achieve both high control accuracy and natural motion synthesis, effectively balancing the two objectives.
Although our method significantly mitigates the trade-off between motion realism and control fidelity, there are still failure cases. For instance, when provided with an extreme or out-of-distribution trajectory (e.g., a pelvis moving at a height of 3 meters above ground), the model still generates a walking motion on the ground. We attribute this to the fact that such a trajectory is far outside the training distribution. However, this issue could be easily addressed by relatively representing joint heights rather than absolute coordinates.

\section{Conclusion and Limitations}

In conclusion, we propose CMC, a decoupled framework to coordinate text and trajectory conditions for trajectory-controlled motion generation. The first stage generates the trajectories of the pelvis and controlled joints based on the given trajectory, coupled with textual descriptions. The second stage generates the full-body motion based on the partial observations from the first stage. Experiments on HumanML3D and KIT-ML datasets validate the effectiveness of CMC. Although CMC exactly tackles the conflicts in text and trajectory-controlled motion generation, it still has limitations of imperfect trajectory error and location error. The reason is that trajectory guidance is applied in the raw motion space instead of the latent space, which leads to nonuniform gradient accumulation in sequence-like data. This motivates us to develop a better framework for better trajectory control.

{
    
    \bibliographystyle{plain}
    \bibliography{refs}
}

\end{CJK}
\end{document}